\documentclass{article}

\usepackage[preprint]{colm2026_conference}

\usepackage{microtype}
\usepackage{hyperref}
\usepackage{url}
\usepackage{booktabs}
\usepackage{amsmath,amssymb}
\usepackage{graphicx}
\graphicspath{{../figures/mac/}{../figures/latent_truth/}{../figures/paraphrase/}{../figures/intervention/}{figures/}}
\usepackage{multirow}
\usepackage[table]{xcolor}
\usepackage{cleveref}
\usepackage{subcaption}
\usepackage{tikz}
\usetikzlibrary{arrows.meta, positioning, decorations.pathreplacing, calc}
\usepackage{float}
\usepackage{float}
\usepackage[inline]{enumitem}
\usepackage{lineno}
\usepackage{amsmath}
\DeclareMathOperator{\logit}{logit}
\definecolor{darkblue}{rgb}{0, 0, 0.5}
\hypersetup{colorlinks=true, citecolor=darkblue, linkcolor=darkblue, urlcolor=darkblue}

\definecolor{visual}{RGB}{52,120,198}
\definecolor{prior}{RGB}{214,137,16}

\title{Arbitration Failure, Not Perceptual Blindness: \\ How Vision-Language Models Resolve Visual-Linguistic Conflicts }

\newcommand{\aspace}{\hspace{0em}}
\newcommand{\zhaw}{$^{\heartsuit}$}
\newcommand{\oxford}{$^{\spadesuit}$}
\author{%
  Farhad Nooralahzadeh\zhaw, \aspace
  Omid Rohanian\oxford,  \aspace
  \textbf{Yi Zhang}\zhaw,   \aspace 
   \textbf{Jonathan F{\"u}rst}\zhaw,   \aspace
  \textbf{Kurt Stockinger}\zhaw\aspace  \\
  \zhaw{Institute of Computer Science}, Zurich University of Applied Sciences, Switzerland \\
  \oxford{University
of Oxford, UK.}  \aspace\\
  \texttt{farhad.nooralahzadeh@zhaw.ch}
}

\begin{document}

\ifcolmsubmission
\linenumbers
\fi

\maketitle

\begin{abstract}
When a Vision-Language Model (VLM) sees a blue banana and answers "yellow", is the problem of \emph{perception} or \emph{arbitration}?
We explore the question in ten VLMs with various sizes and reveal an \textbf{Encoding--Grounding Dissociation}: models that fail to report what they see  (and thus provide a wrong answer) still encode the visual evidence as strongly as models that provide the correct answer.
Using \textbf{Multimodal Arbitration Crossover} (MAC) analysis with layer-by-layer Logit Lens probing, we track the competition between visual and prior signals across every layer of each model.
We show that visual attributes can be linearly decodable from early layers (AUC\,$>$\,0.86). The accuracy remains nearly identical for both successful and failed samples. However, the gap in the final-layer logit---not the strength of encoding---better predicts grounding outcomes with a correlation of $\rho = 0.847$.
After having studied 
when VLMs base their answers on image clues rather than prior knowledge, we want to understand the causal relationships. We establish causality through \textbf{full-sequence activation patching}. The standard last-token interventions in LLM interpretability do not affect VLMs. In contrast, replacing the full token sequence at layers identified by MAC alters 60 to 84\% of outputs.
Partial-token decomposition shows that image tokens carry almost all of the causal impact, while text tokens have none. Scaling addresses the remaining architectural differences to achieve perfect retention.
Moving \textbf{from diagnosis to intervention}, we show that training-free activation steering---both linear and sparse autoencoder-guided---in early layers can improve visual grounding by up to +3.8\% with degrading performance in some setups.
Overall, these findings lead to a clear conclusion: VLMs already see well, but the challenge is acting on what they see. Targeted interventions can help to bridge this gap.
\end{abstract}
\begin{figure}[H]
\centering
\scalebox{0.9}{%
\begin{tikzpicture}[
    >=Stealth,
    font=\scriptsize,
    box/.style={draw, rounded corners=3pt, thick, align=center, inner sep=2pt},
    phase/.style={draw, rounded corners=2pt, align=center, inner sep=3pt, font=\scriptsize},
]
\node[box, fill=blue!5, inner sep=3pt] (input) {%
    \includegraphics[height=1.1cm]{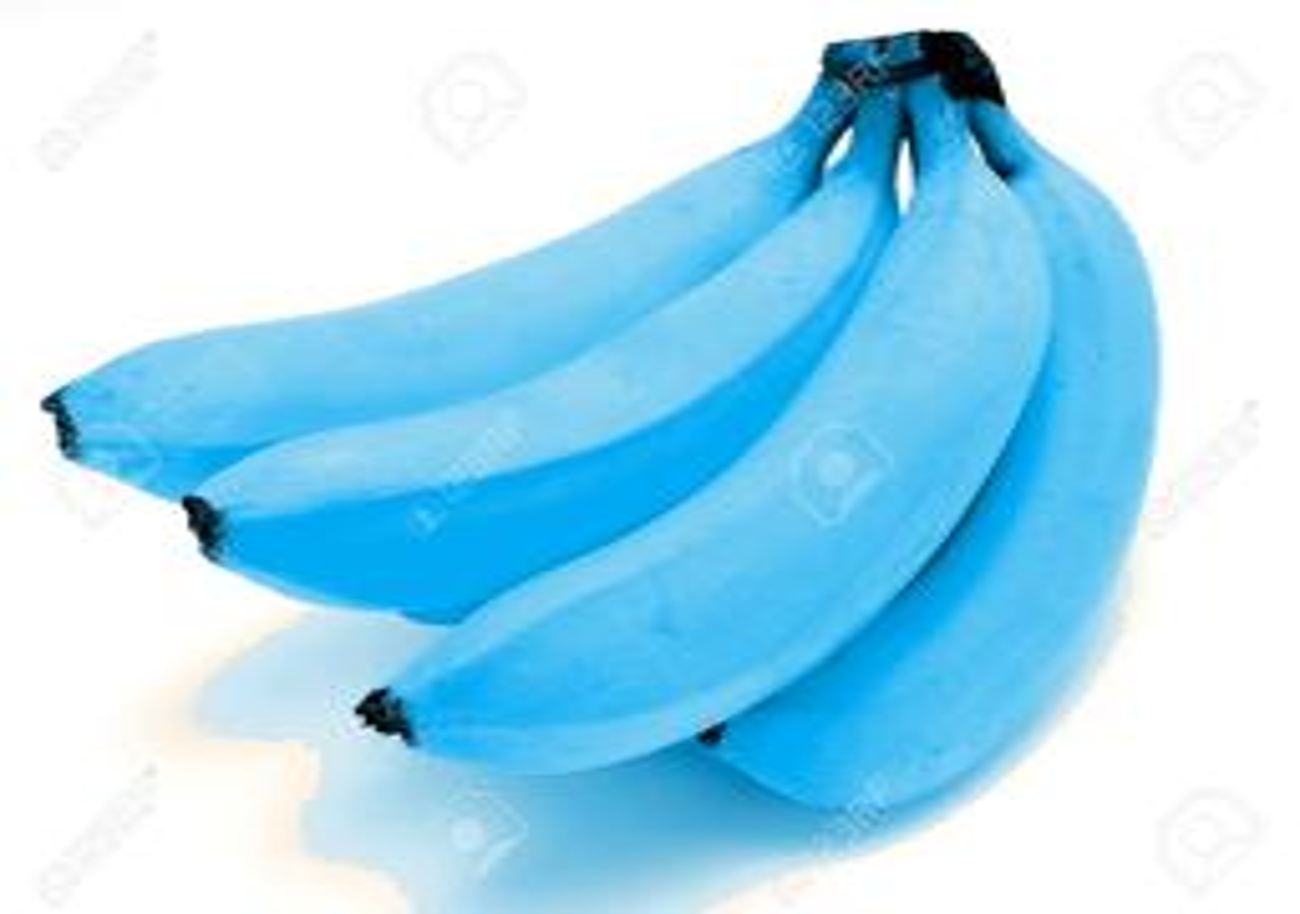}\\[-1pt]
    {\tiny\itshape ``What color is the banana?''}%
};
\node[above=1pt of input, font=\tiny\bfseries] {Input};

\node[phase, fill=green!10, minimum width=2.4cm, minimum height=0.8cm,
      right=0.5cm of input] (enc) {%
    {\tiny\bfseries Perception}\\[1pt]
    encodes {\color{blue!70!black}\textbf{``blue''}} {\color{green!50!black}\checkmark}%
};

\node[phase, fill=red!8, minimum width=2.4cm, minimum height=0.8cm,
      right=0.3cm of enc] (arb) {%
    {\tiny\bfseries Arbitration}\\[1pt]
    prior {\color{prior}\textbf{``yellow''}} wins {\color{red!70!black}$\boldsymbol{\times}$}%
};

\draw[decorate, decoration={brace, amplitude=3pt, raise=2pt}]
    (enc.north west) -- (arb.north east)
    node[midway, above=5pt, font=\tiny\bfseries] {VLM Internal Processing};

\node[box, fill=orange!8, minimum width=1.8cm, minimum height=1cm,
      right=0.5cm of arb] (out) {%
    {\footnotesize\color{prior}\textbf{``yellow''}}\\[1pt]
    {\tiny\color{red!70!black}$\boldsymbol{\times}$ \normalcolor\itshape incorrect}%
};
\node[above=1pt of out, font=\tiny\bfseries] {Output};

\draw[->, thick, blue!40!black] (input) -- (enc);
\draw[->, thick, gray!50] (enc) -- (arb);
\draw[->, thick, red!40!black] (arb) -- (out);

\end{tikzpicture}%
}
\caption{When shown a blue banana, VLMs correctly \emph{perceive} and encode the visual signal ("blue") in their hidden states. However, the downstream \emph{arbitration} mechanism overrides the correctly encoded percept by the linguistic prior ("yellow"). The bottleneck is not perceptual blindness but arbitration.}
\label{fig:teaser}
\end{figure}
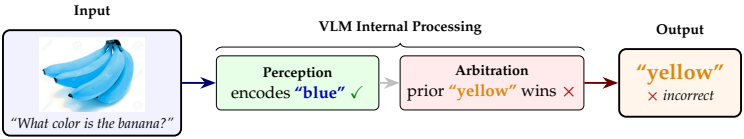

\section{Introduction}
\label{sec:introduction}

Vision-language models (VLMs) such as LLaVA~\citep{NEURIPS2023_6dcf277e}, Qwen2-VL~\citep{qwen2vl}, and InternVL~\citep{chen2024internvl} perform well on many multimodal benchmarks. However, they often fail when visual evidence disagrees with strong language biases. For instance, if a VLM sees a blue banana and is asked about its color, it might respond "yellow". Understanding this behavior matters in high-stakes situations where users expect models to report what they actually see instead of relying on prior assumptions.
A common explanation is \emph{perceptual blindness}. In this scenario, the vision encoder does not capture the important visual detail, so the language part never gets the necessary information. We explore this idea in a study of ten VLMs ranging from 7B to 72B parameters. Contrary to the perceptual-blindness view, we find that the visual information is clearly captured across models, even when the final answer aligns with the language bias. This means that many grounding errors happen despite the model receiving the visual information. The issue lies in how the model utilizes that visual information during generation, not in its ability to capture it.

Our investigation proceeds in four stages:\\
    \textbf{(1) Multimodal Arbitration Crossover (MAC) analysis} (\Cref{sec:mac}). We use a Logit Lens readout with a six-variant token-matching protocol to trace, layer by layer, whether the model prefers the image-consistent answer or the prior-consistent answer. This shows a \emph{crossover point} (the MAC layer) where the visual logit first reliably exceeds the prior logit. The crossover depth varies widely across architectures (36--71\% of layers) and depends on the attribute that is causing the conflict. Scaling shifts the crossover earlier and increases the final-layer visual advantage. For example, InternVL2's \emph{visual-win rate} on Color, which is the fraction of samples where the final-layer logit of the visual tokens exceeds that of the prior token, rises from 58\% (8B) to 87\% (26B).\\
    \textbf{(2) Encoding--grounding dissociation} (\Cref{sec:dissociation}). We examine whether grounding failures occur due to weak encoding. We assess encoding strength in layers before the MAC crossover and compare cases where the model follows the image with those where it follows the prior. The encoding strength is statistically similar in both groups. A second test supports this finding: linear probes extract the visual attribute from early layers (10\% depth; AUC $>$ 0.86) with almost the same accuracy for both groups. In contrast, the results are influenced by a signal further down the line. The logit gap between the final layer's visual and prior signals predicts success ($\rho = 0.847$), while encoding strength does not ($\rho = 0.198$).\\
    \textbf{(3) Causal validation via activation patching} (\Cref{sec:patching}). The first two stages are correlational. To test causality, we use activation patching at the MAC-identified layers. For each example, we inject hidden states from a standard-image run into the counterfactual run and check if the output changes. Full-sequence patching flips 60--84\% of outputs from the visual answer to the prior. In contrast, last-token patching, which is standard in LLM interpretability, only produces a flip rate of 0--1\%. This difference shows that visual information spreads across many image tokens instead of being focused at a single position. A partial-token breakdown indicates that image tokens carry almost all of the causal effect, while patching text tokens yields only a marginal effect. At larger model scales, image-only patching captures the full effect.\\
    \textbf{(4) From diagnosis to intervention} (\Cref{sec:intervention}). Finally, we ask if the diagnostics indicate a way to fix issues at inference time. We apply two training-free steering methods at early layers. The first method is linear activation addition. The second one is sparse autoencoder (SAE) guided steering, which we apply using a residual formulation. Both methods improve visual grounding by as much as +3.8\%. In some configurations, SAE steering method improves grounding, negatively affecting the evaluated examples.

Overall, these findings converge on a single conclusion: When VLMs fail, it's often because of how decisions are made, not because the model can't perceive the information. Targeted, training-free interventions can mitigate this issue to some extent.

\section{Related work}
\label{sec:related}

\textbf{Layerwise probing and Causality tests in VLMs.}
Logit Lens~\citep{nostalgebraist2020logitlens,belrose2023eliciting} enables us to inspect how candidate answers emerge across layers. \citet{golovanevsky-etal-2025-pixels} adapted this to VLMs, but their conclusions depend in part on a narrow token-matching scheme. We revisit that line of work across a broader set of models and with a more conservative matching protocol, then use it to identify where visual and prior-consistent answers begin to diverge.
Probing tells us where visual and prior-consistent answers begin to separate, but not whether those layers actually drive the model's decision. To test that, we use activation patching~\citep{10.5555/3495724.3496763, NEURIPS2021_4f5c422f, 10.5555/3600270.3601532}, which swaps hidden states across inputs to measure causal effects. In text-only LLMs, this is typically done at the last token~\citep{10.5555/3600270.3601532}. In VLMs, however, that setup is a poor fit: visual information is distributed across the image-token sequence rather than concentrated at a single position, and last-token interventions leave 90--98\% of outputs unchanged. This is what motivates our full-sequence patching analysis.

\textbf{VLM interpretability and hallucination.}
Prior mechanistic studies examine only a small number of models~\citep{schwettmann2023multimodal, Palit_2023_ICCV, jiang2025interpreting,DBLP:conf/iclr/NeoO0G0B25}. A related line of work studies object hallucination~\citep{li-etal-2023-evaluating} and methods for mitigating it, including contrastive decoding~\citep{leng2024mitigating}. Our findings are more consistent with an arbitration account of many such failures: visual information may be encoded, yet still lose out to a strong linguistic prior. Recent concurrent work also examines internal multimodal mechanisms more directly, for example by identifying a "fusion band" in a small set of VLMs~\citep{chien2026mint} or pruning hallucination-linked attention heads~\citep{wuliyan2026headlevel}. We complement this literature by tracing visual-prior competition across ten architectures and relating those dynamics to whether the model follows the image or the prior. Standardized benchmarks~\citep{mib-2025} provide a complementary external evaluation perspective.

\textbf{Steering and Sparse Autoencoder.}
Representation engineering~\citep{DBLP:journals/corr/abs-2310-01405} steers LLM behavior by adding contrastive mean directions to hidden states, with applications to truthfulness~\citep{li2023inferencetime} and sentiment~\citep{turner2023steering}. These methods operate on text-only LLMs at a single token position (typically the last). Recent work extends activation steering to VLMs for hallucination mitigation, such as ASD~\citep{su-etal-2025-activation} and SHARP~\citep{wu-etal-2025-sharp}. Our work differs in two respects: (1)~we ground the intervention site in a diagnostic pipeline (MAC analysis + patching) rather than selecting layers heuristically, and (2)~we show that full-sequence steering across all token positions is necessary for VLMs, since visual information is distributed rather than concentrated at the last token.
Sparse autoencoders (SAEs) decompose neural network activations into interpretable, monosemantic features~\citep{bricken2023monosemanticity, huben2024sparse}. \citet{templeton2024scaling} demonstrate that clamping individual SAE features can steer model behavior in text-only LLMs. \citet{pach2025sparse} extend SAEs to VLMs and show that the learned features remain monosemantic in the multimodal setting. We apply SAE-guided steering to visual grounding with two adaptations: bidirectional feature manipulation (amplifying visual, suppressing prior) and a residual application strategy---inspired by residual connections~\citep{he2016deep}---that adds only the decoded delta to the original hidden state, avoiding information loss from lossy SAE reconstructions.
\begin{figure}[t]
\centering
\begin{minipage}[c]{0.3\textwidth}
\centering
\resizebox{\linewidth}{!}{%
\begin{tikzpicture}[
    >=Stealth,
    font=\scriptsize,
    box/.style={draw, rounded corners=2pt, thick, align=center, inner sep=3pt, minimum height=0.9cm},
    exbox/.style={draw, rounded corners=2pt, thick, align=left, inner sep=3pt},
    arr/.style={->, thick},
]
\node[box, fill=gray!10, minimum width=6.2cm] (m1) at (0.0,1.3) {
    \textbf{1) Layer logits}\\
    for each $\ell$: get\\
    $\logit_{v}^{(\ell)},\logit_{p}^{(\ell)}$ (blue vs.\ yellow)
};
\node[box, fill=blue!8, minimum width=6.2cm, below=0.45cm of m1] (m2) {
    \textbf{2) Six-form max}\\
    $\mathcal{T}_{v}$: blue/Blue/BLUE/\_blue/\_Blue/hex\\
    $\mathcal{T}_{p}$: yellow/Yellow/YELLOW/\_yellow/\_Yellow/hex\\
    define $\logit_v^{(\ell)}=\max_{t\in\mathcal{T}_v}\ell_t^{(\ell)}$, \;
    $\logit_p^{(\ell)}=\max_{t\in\mathcal{T}_p}\ell_t^{(\ell)}$
};
\node[box, fill=green!10, minimum width=6.2cm, below=0.45cm of m2] (m3) {
    \textbf{3) Pick MAC}\\
    first stable $\ell^*$ where\\
    $\logit_{v}^{(\ell)}>\logit_{p}^{(\ell)}$ and persists
};
\draw[arr] (m1) -- (m2);
\draw[arr] (m2) -- (m3);
\node[font=\tiny, align=left, below=0.12cm of m3] (m3note) {
check $\ell$ and $\ell{+}1$ to avoid transient crossings
};
\node[exbox, fill=orange!8, minimum width=6.2cm, minimum height=1.2cm, below=0.45cm of m3] (mx) {
\textbf{Banana example}\\
$\ell=12$: \textcolor{blue!70!black}{blue variants} $=\{1.70,1.44,0.93,\ldots\}$\\
\hspace*{1.2em}\textcolor{orange!80!black}{yellow variants} $=\{1.55,1.31,0.88,\ldots\}$\\
\hspace*{1.2em}$\Rightarrow\ \logit_v^{(12)}=\textcolor{blue!70!black}{\mathbf{1.70}},\ \logit_p^{(12)}=\textcolor{orange!80!black}{\mathbf{1.55}}$\\
$\ell=13$: $\logit_v^{(13)}=\textcolor{blue!70!black}{1.88},\ \logit_p^{(13)}=\textcolor{orange!80!black}{1.42}$ (still $v>p$)\\
$\Rightarrow\ \text{first stable crossover } = \text{MAC}=12$
};
\draw[arr, green!50!black] (m3) -- (mx);
\node[font=\tiny, align=center, below=0.14cm of mx] {$v$ = visual token,\;\; $p$ = prior token,\;\; $\ell^*$ = MAC};
\end{tikzpicture}%
}
\end{minipage}\hfill
\begin{minipage}[c]{0.7\textwidth}
\centering
\includegraphics[width=1\textwidth]{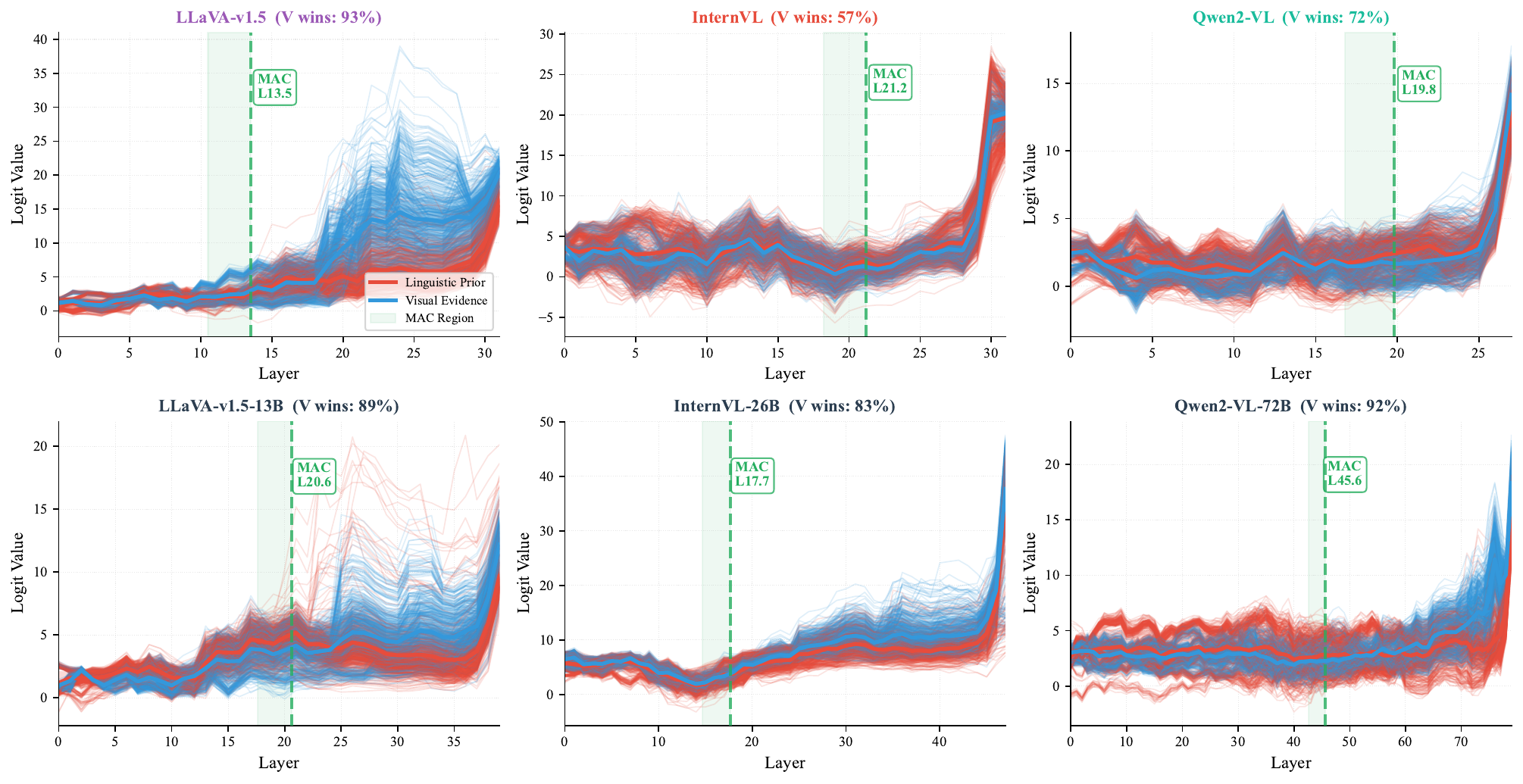}
\end{minipage}

\caption{\textbf{Left:} concrete banana example with the max-variant rule: at each layer, we take the best-matching blue and yellow token forms, then define MAC as the first stable layer where visual ($v$) exceeds prior ($p$), verified by persistence at the next layer. \textbf{Right:} MAC logit trajectories (color, 493 samples). Each subplot shows per-layer visual (blue) and prior (orange) logits with sample traces and mean $\pm$1\,std shading. Green dashed line: MAC crossover. ``V wins'' indicates the percentage of samples where the visual logit exceeds the prior logit at the final layer. \textbf{\small Top:} three primary models (7--8B). \textbf{\small Bottom:} scaled-up variants. See Appendix \Cref{app:mac-figures-full} for all ten models.}
\label{fig:mac-trajectories}
\end{figure}
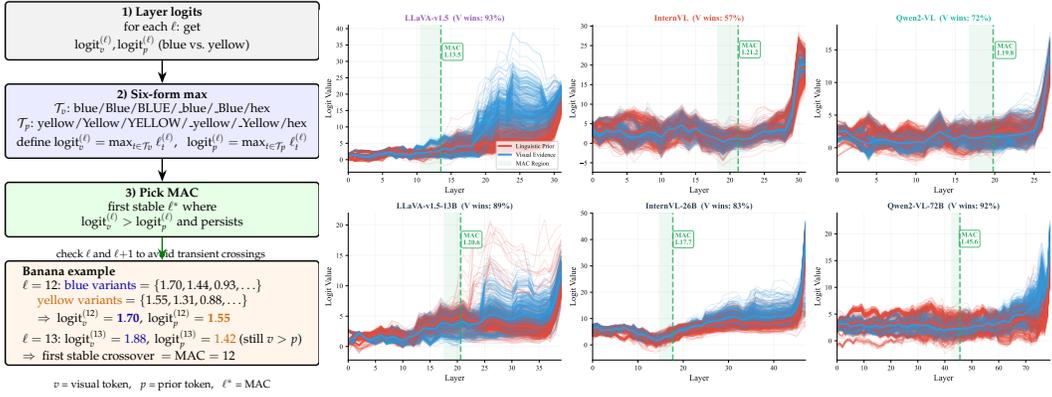

\section{Experimental setup}
\label{sec:setup}
\textbf{Dataset.}
We use the Visual-Counterfact dataset~\citep{golovanevsky-etal-2025-pixels}, which provides images with visually altered properties to create controlled visual-linguistic conflicts. The dataset contains two distinct splits, each corresponding to a specific visual attribute reasoning task: \emph{Color} (493 examples, like a banana that's blue instead of yellow) and \emph{Size} (727 examples, where the usual size relationships between objects are flipped).

\textbf{Models.}
We consider ten VLMs from four architecture families, three vision encoder designs, and with parameters from 7B to 72B. The seven \emph{primary models} at 7 to 8B scale are: LLaVA-v1.5, LLaVA-Next, and LLaVA-OneVision~\citep{NEURIPS2023_6dcf277e} which use CLIP/SigLIP vision encoders with Llama-2, Mistral, and Qwen2 as LLM backbones, respectively; they have  28--32 layers. Next is Qwen2-VL~\citep{qwen2vl} which combines Qwen-ViT and Qwen2-7B and has 28 layers. InternVL2~\citep{chen2024internvl} includes InternViT-6B and InternLM2-8B, featuring 32 layers. BLIP-2 uses EVA-ViT-G and OPT-2.7B with 32 layers. Finally, DeepSeek-Janus-Pro includes SigLIP and DeepSeek-7B, with 30 layers. Three \emph{scaled-up variants} allow for analysis of within-family scaling: LLaVA-v1.5-13B with 40 layers, InternVL2-26B with 48 layers, and Qwen2-VL-72B with 80 layers.



\begin{table}[t]
\centering
\scriptsize
\renewcommand{\arraystretch}{0.85}
\resizebox{\linewidth}{!}{%
\begin{tabular}{lc|ccc|ccc|ccc|c}
\toprule
\textbf{Model} & \textbf{$L$} & \multicolumn{3}{c|}{Color} & \multicolumn{3}{c|}{Size} & \multicolumn{4}{c}{Latent (Color)} \\
\cmidrule(lr){3-5}\cmidrule(lr){6-8}\cmidrule(lr){9-12}
 & & MAC & R\% & D\% & MAC & R\% & D\% & 25\% & 50\% & 75\% & Cos@75 \\
\midrule
\multicolumn{12}{l}{\textit{Primary models (7--8B)}} \\
DeepSeek-Janus & 30 & L10.8 & 91\% & 36\% & L14.6 & 70\% & 49\% & 14.29 & 63.26 & \textbf{178.60} & 0.961 \\
LLaVA-Next & 32 & L12.7 & 82\% & 40\% & L15.0 & 64\% & 47\% & 0.24 & 1.73 & \textbf{3.80} & 0.977 \\
LLaVA-v1.5 & 32 & L13.5 & 91\% & 42\% & L16.9 & 75\% & 53\% & 1.07 & 9.60 & \textbf{31.67} & 0.935 \\
LLaVA-OneVision & 28 & L16.8 & 96\% & 60\% & L17.1 & 92\% & 61\% & 3.80 & 20.81 & \textbf{53.04} & 0.899 \\
Qwen2-VL & 28 & L19.8 & 75\% & 71\% & L17.8 & \textbf{85\%} & 64\% & 5.71 & 12.58 & \textbf{30.85} & 0.967 \\
BLIP-2 & 32 & L19.9 & 92\% & 62\% & L22.4 & 58\% & 70\% & 1.81 & 4.82 & \textbf{27.35} & 0.931 \\
InternVL2 & 32 & L21.2 & 58\% & 66\% & L17.5 & \textbf{76\%} & 55\% & 5.16 & 23.06 & \textbf{44.74} & 0.962 \\
\midrule
\multicolumn{12}{l}{\textit{Scaled-up variants (13--72B)}} \\
InternVL2-26B & 48 & L17.7 & 87\% & 37\% & L26.0 & 74\% & 54\% & 12.44 & 51.95 & \textbf{102.28} & 0.904 \\
LLaVA-v1.5-13B & 40 & L20.6 & 91\% & 52\% & L18.1 & 81\% & 45\% & 2.77 & 18.18 & \textbf{27.32} & 0.978 \\
Qwen2-VL-72B & 80 & L45.6 & \textbf{96\%} & 57\% & --- & --- & --- & 6.95 & 61.08 & \textbf{146.10} & 0.938 \\
\bottomrule
\end{tabular}%
}
\caption{MAC crossover and latent encoding (Color: 493 samples; Size: 727). \textbf{Color} / \textbf{Size}: \emph{MAC} = mean crossover layer; \emph{R\%} = \% samples where visual logit wins at final layer; \emph{D\%} = MAC depth as \% of total layers $L$. \textbf{Latent (Color)}: L2 distance at 25\%, 50\%, and 75\% of mean MAC depth (higher = stronger encoding); \emph{Cos@75} = cosine similarity at 75\% depth. Per-model rankings partially reverse between Color and Size; scaled variants show $2$--$5\times$ stronger L2 at 75\%.}
\label{tab:mac-latent}
\end{table}
\section{Multimodal Arbitration Crossover (MAC) analysis}
\label{sec:mac}

For a VLM with $L$ transformer layers, we extract the last-token hidden state $h^{(\ell)}$ at each layer $\ell$ and project it through the final layer norm and language model head to obtain vocabulary-level logits. Following prior Logit Lens practice~\citep{nostalgebraist2020logitlens,belrose2023eliciting}, $h^{(\ell)}$ is taken from the residual stream at layer $\ell$. Intuitively, this asks: "if decoding stopped at layer $\ell$, which answer token is currently preferred?" We use the last token because the final answer decision in autoregressive VLMs is read out there.  Prior work~\citep{golovneva2024pixels} monitors only two variants per answer (e.g., "blue" and "Blue"), whereas we look at six versions of each visual and prior token: lowercase, capitalized, uppercase, two space-prefixed forms, and hex subwords.
For each of the two competing answers (visual and prior), we track the maximum logit across all six surface-form variants:
\begin{equation} \text{logit}_{\text{visual}}^{(\ell)} = \max_{t \in \mathcal{T}_{\text{visual}}} \left[ \text{LN}(h^{(\ell)}) \cdot W_{\text{lm}}[t] \right], \quad
    \text{logit}_{\text{prior}}^{(\ell)} = \max_{t \in \mathcal{T}_{\text{prior}}} \left[ \text{LN}(h^{(\ell)}) \cdot W_{\text{lm}}[t] \right]
\end{equation}
$\mathcal{T}_{\text{visual}}$ and $\mathcal{T}_{\text{prior}}$ are the token ID sets, LN is the final layer normalization, and $W_{\text{lm}}$ is the language model head. 
As it shown in \Cref{fig:mac-trajectories} (Left block) the \textbf{MAC layer} for a given sample is then defined as the first layer at which the visual logit stably exceeds the prior logit, subject to persistence and transition constraints (see \Cref{app:implementation} for details).


\textbf{Crossover dynamics across architectures.}
\Cref{fig:mac-trajectories} shows the logit competition among the three primary models together with their larger variants. On counterfactual samples, a prediction is counted as correct only if the visual token still wins at the final layer; if the prior token wins (e.g., blue banana → "yellow"), that sample is a failure and contributes to lower final-win rates. The models behave differently. Among small models, LLaVA-v1.5 responds to the visual signal early, while models such as (InternVL2, Qwen2-VL) wait until the final third to cross over and keep narrow margins throughout. Trajectories for all seven primary models are in \Cref{app:mac-figures-full}. \Cref{tab:mac-latent} presents data to illustrate this range. It evaluates each model across all counterfactual examples: color: 493, and size: 727.

\begin{table}[t]
\centering

\scriptsize
\renewcommand{\arraystretch}{0.85}
\resizebox{\linewidth}{!}{%
\begin{tabular}{l|cc|cc|cc|ccc|ccc|c}
\toprule
\textbf{Model} & \multicolumn{2}{c|}{VISUAL} & \multicolumn{2}{c|}{PRIOR} & \multicolumn{2}{c|}{} & \multicolumn{3}{c|}{Full-seq.\ patch} & \multicolumn{4}{c}{Token-type} \\
\cmidrule(lr){2-3}\cmidrule(lr){4-5}\cmidrule(lr){6-7}\cmidrule(lr){8-10}\cmidrule(lr){11-14}
 & L2 & $n$ & L2 & $n$ & Ratio & $p$(MW) & Chg & Flip & Cond. & Full & Img & Txt & Ret. \\
\midrule
\multicolumn{14}{l}{\textit{Primary models (7--8B)}} \\
DeepSeek-Janus & 178.84 & 434 & 176.85 & 59 & 1.01$\times$ & 0.946 & 99 & \textbf{82} & 94 & 82 & \textbf{83} & 0 & 101\% \\
LLaVA-Next & 3.82 & 357 & 3.75 & 136 & 1.02$\times$ & 0.559 & 96 & \textbf{60} & 82 & 60 & \textbf{58} & 0 & 97\% \\
LLaVA-v1.5 & 32.02 & 460 & 26.75 & 33 & 1.20$\times$ & 0.002 & 99 & \textbf{78} & 94 & 78 & \textbf{77} & 1 & 99\% \\
LLaVA-OneVision & 53.04 & 471 & 53.20 & 22 & 1.00$\times$ & 0.916 & 98 & \textbf{68} & 88 & 68 & \textbf{67} & 0 & 99\% \\
Qwen2-VL & 30.98 & 356 & 30.50 & 137 & 1.02$\times$ & 0.347 & 98 & \textbf{82} & 93 & 82 & \textbf{57} & 0 & 70\% \\
BLIP-2 & 27.50 & 461 & 25.23 & 32 & 1.09$\times$ & 0.093 & \multicolumn{7}{c}{---} \\
InternVL2 & 43.97 & 281 & 45.76 & 212 & 0.96$\times$ & 0.049 & 97 & \textbf{63} & 79 & 63 & \textbf{51} & 1 & 81\% \\
\midrule
\multicolumn{14}{l}{\textit{Scaled-up variants (13--72B)}} \\
InternVL2-26B & 102.17 & 411 & 102.84 & 82 & 0.99$\times$ & 0.776 & 98 & \textbf{72} & 86 & 72 & \textbf{72} & 2 & 100\% \\
LLaVA-v1.5-13B & 26.60 & 437 & 32.93 & 56 & 0.81$\times$ & 0.017 & 98 & \textbf{73} & 88 & 73 & \textbf{73} & 0 & 100\% \\
Qwen2-VL-72B & 145.61 & 454 & 151.87 & 39 & 0.96$\times$ & 0.042 & 100 & \textbf{84} & 92 & 84 & \textbf{84} & 0 & 100\% \\
\bottomrule
\end{tabular}%
}
\caption{\textbf{Left:} encoding at 75\% MAC depth (Color, 493 samples per model): L2 between counterfactual vs.\ standard hidden states for samples where the model answered with the visual vs.\ prior token; \textbf{Ratio} = VISUAL/PRIOR mean L2; $p$(MW) = Mann--Whitney $p$. Ratio $\approx 1$ favors arbitration failure over perceptual blindness. \textbf{Right:} activation patching at the MAC probe layer (100 Color samples per model): \textbf{Chg}/\textbf{Flip}/\textbf{Cond.} = \% any change, \% visual$\rightarrow$prior flips, \% flips conditioned on baseline-visual; \textbf{Full}/\textbf{Img}/\textbf{Txt} = flip counts; \textbf{Ret.}\ = \textbf{Img}/\textbf{Full}. BLIP-2: patching not run (---). Last-token patching 0--1\% flip; see \Cref{app:patching-extra}.}
\label{tab:encoding-patching}
\end{table}

DeepSeek-Janus starts to resolve the conflict at 36\% depth, which is the first third of the network. Qwen2-VL waits until 71\%. However, speed alone, does not predict reliability well. LLaVA-OneVision has the highest color success rate at 96\% with a crossover at 60\% depth, while InternVL2 only achieves 58\% at a similar 66\%. The position of the crossover is less important than how wide the logit gap is. The final-layer logit landscape illustrates this clearly. LLaVA-OneVision and BLIP-2 rank the visual token at positions 2 and 3, with gaps of +3.2 to +5.7 logits. This leaves little room for the earlier result to regain the top position. InternVL2 and LLaVA-Next depend on margins of +0.6 to +1.0 at ranks 69 to 559. In this scenario, small changes can affect the outcome. Although there are only seven primary models, this finding is preliminary, but it indicates a clear trend: the crossover depth is less important than the confidence the model shows once it reaches that point.

\textbf{Attribute-specific crossover: color vs.\ size.}
\label{sec:mac-size}
The crossover is also attribute-dependent. The average depths are similar for color and size (54\% vs.\ 57\%), but the ranking per-model shift. Qwen2-VL and InternVL2, which are the slowest for color at 71\% and 66\%, drop to 64\%  and 55\% for size, respectively. This suggests that there are \emph{attribute-specific arbitration pathways} rather than a single universal mechanism. We note, however, that size comparisons involve comparative reasoning ("Which is bigger?") rather than simple recognizing an attribute. Altering the question to ("Which is \emph{smaller}?") drops agreement to 48--54\% (\Cref{app:paraphrase}), which hints at keyword matching, so we treat size rates with more caution throughout.

\textbf{Paraphrase robustness.}
To eliminate phrasing artifacts, we test eight different question variations on three main models (100 samples each). The logit trajectories show a correlation of $r = 0.958$ across phrasings. The winning models agree 93.9\% of the time, and 81.7\% of samples remain perfectly stable across all eight variations. Qwen2-VL-72B demonstrates similar stability (mean 93\%); InternVL2-26B is more sensitive (mean 80\%), indicating that scaling does not always improve robustness to syntactic variation (\Cref{app:paraphrase}).

\textbf{Model scaling analysis.}
\label{sec:mac-scaling}
The bottom rows of \Cref{tab:mac-latent} and the bottom panel of \Cref{fig:mac-trajectories} illustrate how scaling affects these dynamics. The weakest models benefit the most: InternVL2 increases from 58\% (8B) to 87\% (26B), while Qwen2-VL goes from 75\% (7B) to 96\% (72B). LLaVA-v1.5, already at 91\%, shows no change. Crossovers also happen earlier as the scale increases (InternVL2-26B: 37\% depth compared to 66\% at 8B; Qwen2-VL-72B: 57\% compared to 71\%), and the logit gaps widen (\Cref{fig:mac-trajectories}, bottom). Essentially, larger models commit to the visual answer earlier and with bigger margins. The crossover, the fast/slow integrator spectrum, and the attribute specificity are consistent across the entire 27$\times$ parameter range.

\section{The Encoding--Grounding Dissociation}
\label{sec:dissociation}

The MAC analysis identifies \emph{where} visual evidence overtakes linguistic priors but does not clarify between two possible failure modes: either the visual signal was never encoded well (a perceptual issue), or it was encoded but later overshadowed by the prior (an arbitration issue). We now test which explanation is correct.

\textbf{Latent truth: all models encode visual information.}
\label{sec:latent-truth}
For each sample and model, we extract hidden states at 25\%, 50\%, and 75\% of the MAC layer for both the counterfactual image ($I_{\text{cf}}$, e.g., a blue banana) and the standard image ($I_{\text{std}}$, e.g., a yellow banana). The L2 distance {\scriptsize $\|h_{\text{cf}}^{(\ell)} - h_{\text{std}}^{(\ell)}\|_2$} measures how much the model's internal representation differentiates between the two images at each depth: larger distances indicate that the model has noted the visual difference, even before producing the output.

\Cref{tab:mac-latent} (right block) confirms that \textbf{every model} encodes visual information before the MAC crossover, with L2 increasing steadily from 25\% to 75\% depth. The absolute values range from 47$\times$ range (DeepSeek-Janus: 178.60 vs.\ LLaVA-Next: 3.80), but the steady pattern is consistent. Notably, Qwen2-VL ranks 5th of 7, which contradicts prior claims~\citep{golovanevsky-etal-2025-pixels}  of a "perceptual blind spot" in that model. Scaling further enhances encoding: Qwen2-VL-72B reaches 146.10 (4.7$\times$ the 7B variant). Detailed probing at 20 depth points (\Cref{fig:lt-fine}) shows that encoding increases steadily to the final layer and is never suppressed.

\textbf{Failure analysis: arbitration, not blindness.}
\label{sec:failure-analysis}
We now ask whether models encode visual information \emph{less} in samples where grounding fails. If the failure was perceptual, L2 distances should be significantly smaller for incorrect samples. 
\Cref{tab:encoding-patching} (left block) compares encoding strength at 75\% MAC depth between the success (VISUAL) and failure (PRIOR) groups.

{The success/failure L2 ratio ranges from $[0.81, 1.20]$ for all ten models.} When a model answers "yellow" for a blue banana, its hidden state encoded "blue" with almost the same strength as in correct instances. In the most extreme case, LLaVA-v1.5-13B, the ratio is 0.81$\times$, indicating failures encode the visual attribute \emph{more} strongly than successes. Applying Mann-Whitney U tests \citep{doi:https://doi.org/10.1002/9780470479216.corpsy0524}, we show that the two distributions are indistinguishable for six of ten models ($p > 0.05$). Among the four that show significance, three have L2 for the failure group \emph{higher} than the success group (ratios $\leq$\,0.96), which goes against any encoding-deficit explanation.

Considering Linear probe validation,
L2 distance is an overall measure and might not capture the specific attribute information that matters. Therefore, we train logistic regression probes (5-fold CV) at ten evenly spaced layers for each model to classify the visual attribute directly (\Cref{tab:linear-probe} in the appendix). Probes achieve AUC\,$>$\,0.86 at just 10\% depth for every model and exceed 0.99 by mid-depth; visual attributes are \emph{linearly separable} from early on. Probe confidence on success and failure samples is almost identical ($|\Delta| < 0.03$, sign inconsistent across models), indicating that a classifier decodes "blue" with equal accuracy, regardless of whether the model finally outputs "blue" or "yellow". Using encoding strength as a predictor of grounding outcome yields AUC\,=\,0.528 (chance\,=\,0.5), suggesting that encoding carries no information about individual outcomes.

\textbf{What predicts success?}
\label{sec:cross-stage}
Correlating model-level metrics with success rate across the seven main models (Spearman $\rho$; \Cref{fig:cross-stage}) reveals a double dissociation. Encoding strength shows no link to success ($\rho = 0.198$, $p = 0.670$), while the logit gap at the final layer is strongly predictive ($\rho = 0.847$, $p = 0.016$). Encoding does not even predict the logit gap ($\rho = 0.464$, $p = 0.294$); the two metrics are largely independent.
InternVL2 and BLIP-2 illustrate this clearly. InternVL2 encodes more strongly (L2\,=\,44.74) but only succeeds on 58\% of samples; BLIP-2 encodes less strongly (L2\,=\,27.35) yet achieves 92\%, because it maintains a larger logit gap at the output. The model-level sample is small ($n = 7$), but the analysis at the same level (3,451 model-sample pairs, AUC\,=\,0.528) confirms that encoding provides no information about individual outcomes. Qwen2-VL-72B, despite resulting L2\,=\,146.10 (the highest in our study), still fails on 4\% of samples.

We now have clear clues about when a model is likely to answer correctly. A model is usually correct when the visual answer clearly wins \emph{at the final layer}. Strong internal encoding alone does not predict success; what matters is whether the model finally chooses the visual token over the prior token.

\begin{figure}[t]
\centering
\scalebox{0.88}{%
\begin{tikzpicture}[
    >=Stealth,
    font=\scriptsize,
    box/.style={draw, rounded corners=2pt, thick, align=center, inner sep=2pt, minimum height=0.6cm},
    imgbox/.style={draw, rounded corners=2pt, inner sep=2pt, align=center},
    result/.style={draw, rounded corners=2pt, thick, inner sep=2pt, align=center, minimum height=0.55cm},
    arr/.style={->, thick},
    darr/.style={->, thick, dashed, red!60!black},
]
\node[font=\scriptsize\bfseries] at (1.8, 1.55) {(a) Patching procedure};

\node[imgbox, fill=blue!5] (cf_img) at (0, 0.6) {%
    \includegraphics[height=0.65cm]{blue_banana.png}%
};
\node[below=-1pt of cf_img, font=\tiny\itshape] {counterfact.};

\node[box, fill=blue!8, minimum width=1.2cm] (vlm1) at (2.0, 0.6) {%
    {\tiny\bfseries VLM}\\[-2pt]
    {\tiny L1\ldots$L$}%
};

\node[result, fill=blue!10] (out1) at (3.6, 0.6) {%
    {\tiny\color{blue!70!black}\textbf{``blue''}}%
};

\draw[arr, blue!50!black] (cf_img) -- (vlm1);
\draw[arr, blue!50!black] (vlm1) -- (out1);

\node[imgbox, fill=orange!5] (std_img) at (0, -0.7) {%
    \includegraphics[height=0.65cm]{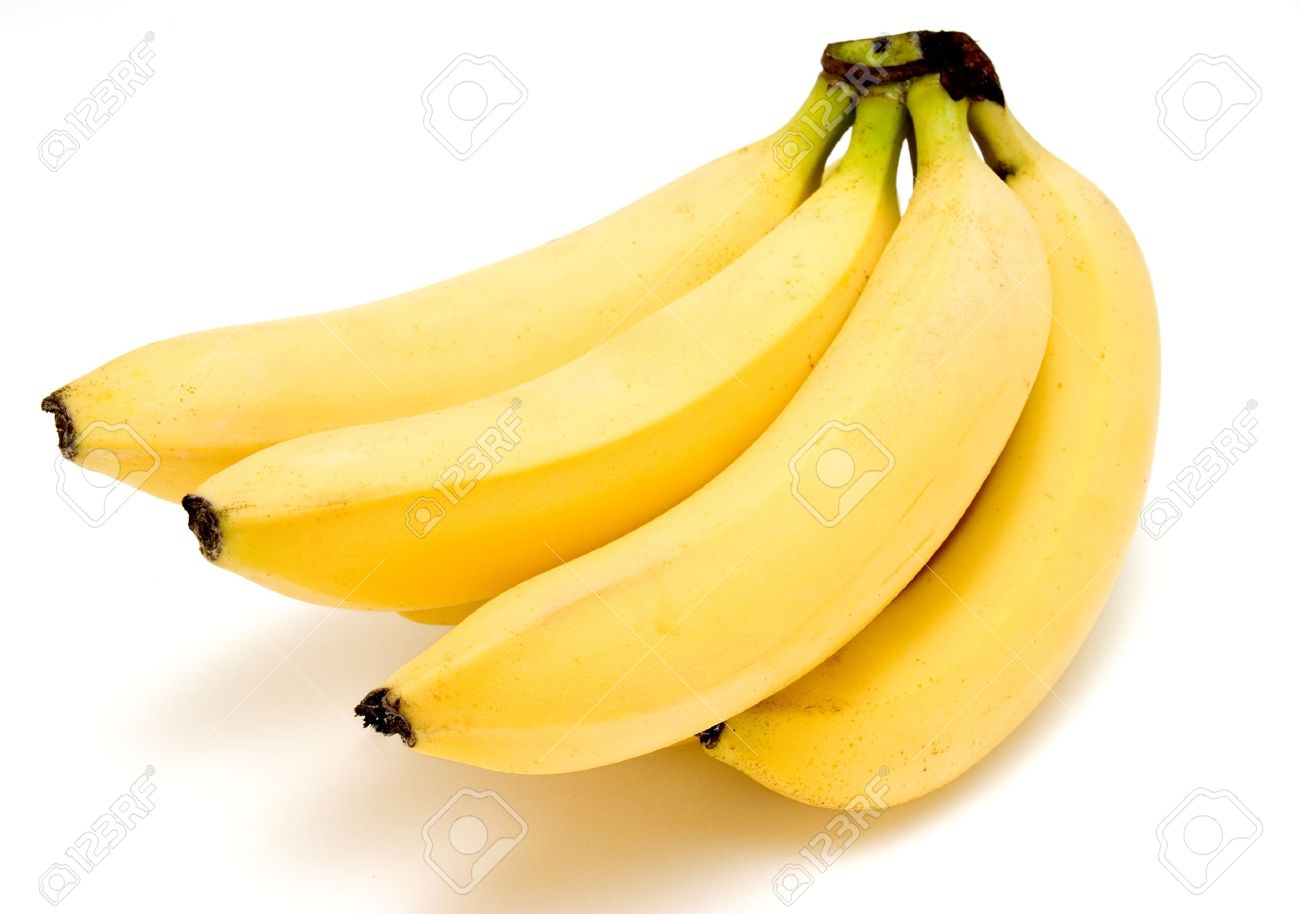}%
};
\node[below=-1pt of std_img, font=\tiny\itshape] {standard};

\node[box, fill=orange!8, minimum width=1.2cm] (vlm2) at (2.0, -0.7) {%
    {\tiny\bfseries VLM}\\[-2pt]
    {\tiny L1\ldots$L$}%
};

\draw[arr, orange!50!black] (std_img) -- (vlm2);

\node[draw, fill=red!10, rounded corners=2pt, inner sep=2pt, font=\tiny, align=center] (hstates) at (3.6, -0.7) {%
    $h^{(\ell^*)}_{\text{std}}$%
};

\draw[arr, orange!50!black] (vlm2) -- (hstates);

\draw[darr, line width=0.8pt] (hstates.north) -- ++(0, 0.35) -| node[pos=0.25, above, font=\tiny\bfseries, red!60!black] {inject} (vlm1.south);

\node[result, fill=orange!10] (out2) at (4.8, 0.6) {%
    {\tiny\color{prior}\textbf{``yellow''}}%
};
\node[above=-1pt of out2, font=\tiny\bfseries, red!60!black] {flip!};

\draw[arr, red!40!black, dashed] (out1) -- (out2);

\draw[gray!40, thick] (5.55, 1.45) -- (5.55, -1.45);

\node[font=\scriptsize\bfseries] at (8.2, 1.55) {(b) Patching scope};

\node[anchor=west, font=\tiny\bfseries] at (5.85, 0.85) {Last-token:};
\foreach \i in {0,...,3} {
    \pgfmathsetmacro{\xpos}{0.32*\i + 7.25}
    \node[draw, fill=blue!12, minimum width=0.28cm, minimum height=0.38cm, inner sep=0pt] at (\xpos, 0.85) {};
}
\node[font=\tiny, inner sep=0pt] at (8.60, 0.85) {$\cdots$};
\node[draw, fill=blue!12, minimum width=0.28cm, minimum height=0.38cm, inner sep=0pt] at (8.90, 0.85) {};
\node[draw, fill=gray!12, minimum width=0.28cm, minimum height=0.38cm, inner sep=0pt] at (9.22, 0.85) {};
\node[draw, fill=red!25, draw=red!60, thick, minimum width=0.28cm, minimum height=0.38cm, inner sep=0pt] at (9.54, 0.85) {};
\node[font=\tiny, anchor=west, align=left] at (9.80, 0.85) {\textbf{0--1\%}\\[-2pt]{\color{red!70!black}$\boldsymbol{\times}$\,\tiny fails}};

\node[anchor=west, font=\tiny\bfseries] at (5.85, -0.05) {Full-seq:};
\foreach \i in {0,...,3} {
    \pgfmathsetmacro{\xpos}{0.32*\i + 7.25}
    \node[draw, fill=red!25, draw=red!60, thick, minimum width=0.28cm, minimum height=0.38cm, inner sep=0pt] at (\xpos, -0.05) {};
}
\node[font=\tiny, inner sep=0pt] at (8.60, -0.05) {$\cdots$};
\node[draw, fill=red!25, draw=red!60, thick, minimum width=0.28cm, minimum height=0.38cm, inner sep=0pt] at (8.90, -0.05) {};
\node[draw, fill=red!25, draw=red!60, thick, minimum width=0.28cm, minimum height=0.38cm, inner sep=0pt] at (9.22, -0.05) {};
\node[draw, fill=red!25, draw=red!60, thick, minimum width=0.28cm, minimum height=0.38cm, inner sep=0pt] at (9.54, -0.05) {};
\node[font=\tiny, anchor=west, align=left] at (9.80, -0.05) {\textbf{63--84\%}\\[-2pt]{\color{green!50!black}\checkmark\,\tiny works}};

\draw[decorate, decoration={brace, amplitude=2pt, mirror, raise=3pt}]
    (6.86, -0.3) -- (9.43, -0.3)
    node[midway, below=5pt, font=\tiny, align=center] {visual info across \emph{all} tokens};

\node[draw, fill=blue!12, minimum width=0.22cm, minimum height=0.22cm, inner sep=0pt] at (6.2, -1.15) {};
\node[font=\tiny, anchor=west] at (6.4, -1.15) {original};
\node[draw, fill=red!25, draw=red!60, thick, minimum width=0.22cm, minimum height=0.22cm, inner sep=0pt] at (7.4, -1.15) {};
\node[font=\tiny, anchor=west] at (7.6, -1.15) {patched from std};

\end{tikzpicture}%
}
\caption{Full-sequence activation patching.
\textbf{(a)}~Run the VLM on a counterfactual image (baseline: “blue”), then inject probe-layer states 
{\scriptsize $h^{(\ell^*)}_{\text{std}}$}
 from a standard-image run. If the output flips to “yellow,” the probe layer causally mediates grounding.
\textbf{(b)}~Last-token patching fails; full-sequence patching succeeds, reflecting that VLMs spread visual information across all tokens.
}
\label{fig:patching-schematic}
\end{figure}

\section{Causal validation via full-sequence activation patching}
\label{sec:patching}

In the previous sections, we show \emph{where} visual evidence surpasses prior information through MAC analysis. We further demonstrate that visual information \emph{exists} in hidden states no matter the grounding outcome, revealed through latent truth probing. However, correlation is not causation: the layers identified by MAC could simply reflect a crossover without actually affecting the model's decision. We address this issue with activation patching, which measures the \emph{indirect effect}~\citep{10.5555/3495724.3496763} of the MAC probe layer on the output.
The process is shown in \Cref{fig:patching-schematic}. For each sample, we run the model on both the counterfactual image (e.g., a blue banana) and the standard image (a yellow banana). We capture hidden states from the standard run at the MAC probe layer, and then re-run the counterfactual forward pass with those states injected. If the output shifts from the visual answer to the prior, the probe layer \emph{causally affects} the grounding decision. In LLM interpretability, it is typical to patch only the last-token position~\citep{10.5555/3600270.3601532}; we find that this approach does not work for VLMs because visual information spreads across many image tokens.

\Cref{tab:encoding-patching} (right block)  presents patching results for nine models (100 samples each). Last-token patching, which is the usual intervention method in LLM interpretability~\citep{10.5555/3600270.3601532}, keeps 90 to 98\% of VLM outputs unchanged (detailed results in \Cref{app:patching-extra}). This is important to point out: the technique that effectively localizes factual recall in text-only models fails in VLMs, because visual information is distributed across the entire token sequence instead of being focused at one point.
Full-sequence patching leads to very different results. Replacing hidden states at all token positions shifts 60 to 84\% of outputs from the visual answer to the prior, with 79 to 94\% based on baseline-visual samples. Notably, there are \emph{zero reverse flips} across all 900 patched samples. The intervention can push the model away from the visual answer but never toward it, confirming unidirectional causal mediation at the MAC layer. Non-flip cases (16--40\%) arise when the baseline already followed the prior or when the model switched to a third token, suggesting some redundancy across layers.

\textbf{Token-type decomposition.}
The token-type columns of \Cref{tab:encoding-patching} break down patching by token type. Text-only patching results in  0 to 2\% flips across all nine models; text tokens do not provide any causal information regarding the grounding decision. Image-only patching captures most of the effect, although retention differs by architecture: the LLaVA family and DeepSeek-Janus retain 97 to 101\%, while Qwen2-VL retains only 70\%, suggesting that its architecture shifts some visual information into text-token positions. Slightly exceeding 100\% (observed only for DeepSeek-Janus) arises from sampling variance in the 100-sample evaluation: in rare cases, the unpatched text tokens introduce minor interference, making image-only patching flip one additional sample. This difference disappears with the larger models. \emph{All three scaled-up variants achieve 100\% retention}, meaning that with sufficient model size, patching image tokens alone recovers the full causal effect.

\section{From diagnosis to intervention}
\label{sec:intervention}

The previous sections show that VLMs encode visual information correctly but do not act on it. The arbitration issue is linked to the layers identified by MAC. A natural question is whether these diagnostic findings lead to a practical solution. We test two lightweight, inference-time steering methods that require no fine-tuning: \emph{linear steering} and \emph{SAE-guided steering}. We focus on three models, namely: LLaVA-v1.5-7B, InternVL2-8B, and Qwen2-VL-7B, because they represent three distinct architecture families (CLIP/Vicuna, InternViT/InternLM, Qwen-ViT/Qwen2), and span the widest range of baseline color accuracy among the 7--8B models (83.6\%--88.4\%). These models also have scaled-up variants (13B, 26B, 72B), making future comparisons easier. We evaluate results on 293 held-out Color samples (200 training, seed 42).
\begin{enumerate*}[label= (\roman*), itemjoin={{, }}, itemjoin*={{ and }}]
\item \textbf{Linear steering.}
Following the contrastive activation addition paradigm~\citep{DBLP:journals/corr/abs-2310-01405, li2023inferencetime, turner2023steering}, we calculate a steering direction {\scriptsize $\mathbf{d} = \overline{h}^{(\text{cf})} - \overline{h}^{(\text{std})}$}from the mean hidden states of counterfactual and standard images at the target layer. We then add $\alpha \cdot \mathbf{d}$ to \emph{all} token positions during inference. Our earlier result indicates that visual information is distributed throughout the entire sequence, making full-sequence steering crucial. Steering that uses only the last token has no effect, which reflects the 0–1\% flip rate of last token patching in \Cref{sec:patching}.

\item \textbf{SAE-guided steering.}
To examine whether more detailed feature control enhances performance over a single direction, we train a Sparse Autoencoder~\citep{bricken2023monosemanticity, huben2024sparse} (SAE; $4\times$ expansion, ReLU, $\lambda{=}0.04$) on hidden states at the target layer. We score features based on their decoder-weighted differential activation $|\Delta_j| \times \|W_{\text{dec}}[:,j]\|_2$, where $\Delta_j$ is the mean activation difference between counterfactual and standard images~\citep{pach2025sparse}. We steer the top-50 visual and top-50 prior features in both directions.
To avoid reconstruction error from lossy SAE decoding~\citep{bricken2023monosemanticity}, we adapt a residual application strategy. This method inspired by residual connections~\citep{he2016deep}: we decode both the original and modified feature vectors, compute only the delta $\delta = \text{decode}(z') - \text{decode}(z)$, and add it to the original hidden state: $h' = h + \delta$. This approach preserves all information that the SAE dose not capture. The full algorithm is given in \Cref{app:steering-details}.
\end{enumerate*}
By doing layers ablation, a key finding is that the \emph{optimal intervention layer is different from the diagnostic MAC layer}. We test five layers per model (10\%--100\% of MAC depth); results are summarized in \Cref{tab:intervention}.

\begin{table}[t]
\centering
\scriptsize
\renewcommand{\arraystretch}{1}
\begin{tabular}{l|c|ccccc|ccccc}
\toprule
& & \multicolumn{5}{c|}{\textbf{Linear Steering}} & \multicolumn{5}{c}{\textbf{SAE Steering}} \\
\textbf{Model} & \textbf{Base.} & \textbf{Layer} & $\boldsymbol{\Delta}$\textbf{Acc.} & \textbf{Acc.} & \textbf{Imp} & \textbf{Deg} & \textbf{Layer} & $\boldsymbol{\Delta}$\textbf{Acc.} & \textbf{Acc.} & \textbf{Imp} & \textbf{Deg} \\
\midrule
LLaVA-v1.5  & 86.0 & L12 & +1.4 & 87.4 & 6 & 2 & L4  & \textbf{+2.0} & \textbf{88.1} & 6 & \textbf{0} \\
InternVL2   & 83.6 & L3  & +3.4 & 87.0 & 16 & 6 & L3  & \textbf{+3.8} & \textbf{87.4} & 14 & \textbf{3} \\
Qwen2-VL    & 88.4 & L6  & \textbf{+3.4} & \textbf{91.8} & 11 & \textbf{1} & L3  & +2.0 & 90.4 & 8 & 2 \\
\bottomrule
\end{tabular}
\caption{Steering results at each model's best layer (293 eval samples, Color attribute). \emph{Base.}\ and \emph{Acc.}: visual-grounding accuracy (\%) before and after steering. $\Delta$\emph{Acc.}: net accuracy change (pp). \emph{Imp}/\emph{Deg}: number of samples improved/degraded.}
\label{tab:intervention}
\end{table}
Here are our findings:
\begin{enumerate*}[label= (\arabic*), itemjoin={{, }}, itemjoin*={{ and }}]
\item \textbf{Early-layer intervention outperforms MAC-layer intervention.}
InternVL2 shows this most clearly. At its MAC crossover layer (L26, 66\% depth), both methods achieve an improvement of $+0.3$\% or less. At L3 (12\% depth), linear steering reaches +3.4\% while SAE steering achieves +3.8\%. The trend is consistent: L3 (+3.4\%) $\to$ L6 (+2.4\%) $\to$ L13 (+2.4\%) $\to$ L19 (0.0\%) $\to$ L26 (+0.3\%). Steering must happen \emph{before} the arbitration regime forms, when hidden states are still adaptable. The MAC crossover indicates where the decision is \emph{observed}, not where it should be \emph{changed}.

\item \textbf{SAE steering provides higher precision at early layers.}
SAE steering results in fewer degradations than linear steering while delivering comparable or greater net improvements. LLaVA-v1.5 at L4, SAE improves 6 samples with \emph{no} degradation ($\infty$ precision), while linear steering at L12 improves 6 but degraded 2 (3:1). InternVL2 at L3, SAE achieves a precision of 4.7:1(14:3) compared to linear's 2.7:1 (16:6). The SAE's advantage comes from its ability to target the distributed visual subspace through 50 independent feature directions rather than one single push. For Qwen2-VL, which has a wider logit gap (\Cref{sec:cross-stage}), linear steering is sufficient and outperforms SAE at every layer.

\item \textbf{All interventions yield net benefits and do not require training.}
Every model shows improvement at its optimal configuration (+1.4\% to +3.8\%) with no regression. Steering only requires about 200 forward passes to compute the direction (or to train the SAE) and adds minimal latency at inference—no weight changes, no fine-tuning, and fully reversible.
\end{enumerate*}
\section{Discussion and Conclusion}
\label{sec:discussion}

Our results make the diagnostic pipeline \emph{actionable}: MAC identifies where visual and prior signals diverge and early-layer intervention improves grounding by +1.4\% to +3.8\% without any training. 
Methodologically, Last-token patching from text-only interpretability ~\citep{10.5555/3600270.3601532} fails for VLMs (0--1\% flip rate); full-sequence or partial-token interventions are needed instead. Applying Token-type decomposition reveals architectural variation at a small scale (Qwen2-VL: 70\% retention) that disappears with scaling (all large models: 100\%).
The encoding--grounding dissociation holds across four LLM backbones (LLaMA, Qwen, InternLM, Vicuna), three vision encoder families (CLIP, SigLIP, InternViT), and a 27$\times$ parameter range. Scaling strengthens encoding 2-5$\times$ and shifts crossovers earlier, yet the dissociation persists even at 72B. This points to a structural feature of current connector-based VLM model architectures rather than a pure capacity limitation. The attribute-specific crossover patterns --change between color and size -- further imply separate arbitration pathway for each attribute type. \\
Our MAC analysis complements MINT~\citep{chien2026mint} by tracking visual versus prior competition across ten architectures and linking it to the encoding--grounding dissociation. Activation steering for reducing hallucinations ~\citep{su-etal-2025-activation, wu-etal-2025-sharp} aligns with our finding that arbitration is the bottleneck, and MAC layers provide a principled way to select intervention sites. Sparse autoencoders~\citep{pach2025sparse} and standardized benchmarks~\citep{mib-2025} offer complementary perspectives.


Across ten vision language models (7B--72B), we establish that grounding failures in visual-linguistic conflicts arise from \emph{arbitration}, not perception. A blue banana is an example to test this: the model captures the idea of "blue" just as well whether it finally says "blue" or "yellow".
MAC analysis shows a universal visual-prior crossover whose depth varies twofold and is attribute-specific. Latent truth probing and linear classifiers show that visual information is faithfully encoded in all models, including failing cases, with encoding strength statistically indistinguishable between success and failure cases; the final-layer logit gap, not encoding, predicts outcomes. Full-sequence activation patching confirms causality: swapping hidden states at MAC-identified layers flips 60--84\% of outputs (vs.\ 0--1\% for last-token patching), with image tokens carrying the full causal signal. Taking the final step from diagnosis to intervention, lightweight activation steering at early layers improves visual grounding by up to +3.8\% with zero degradation in some configurations, confirming that the pipeline produces actionable insights.
The models already see; the challenge is making them act on what they see. Our results show that training-free interventions, guided by the diagnostic pipeline, can begin to close this gap.

\section*{Limitations}
Our analysis relies on synthetic counterfactual images, which provide the controlled conflicts needed to isolate arbitration from confounds but may not capture the full range of visual-linguistic tensions encountered in natural images. Extending to naturally occurring conflicts---unusual real-world colors, ambiguous scenes, fine-grained attributes---is an important direction for future work.

Several design choices bound the strength of our conclusions. The size split shows paraphrase sensitivity (48--54\% agreement under reversed polarity), suggesting partly keyword-driven reasoning and warranting caution when interpreting size grounding rates. InternVL2-26B exhibits lower paraphrase stability (67--92\%) than other models, so its MAC crossover layers may be less robust to syntactic variation, although the baseline phrasing used throughout is among the more stable variants. Causal patching covers nine of ten models with 100 samples each; extending to larger sample sizes would strengthen statistical confidence. The steering experiments (\Cref{sec:intervention}) are limited to the 7--8B primary models; scaling steering to 13B--72B models remains future work. Finally, our primary encoding metric (L2 distance) measures overall representational divergence; the linear probe analysis partially addresses this, but probes trained on specific attribute labels would provide an even more targeted test of whether the visual attribute itself, rather than correlated features, is encoded.



\section*{Ethics statement}
All models and datasets used in this work are publicly available, and the counterfactual images involve synthetic modifications of objects rather than real individuals. Our central finding---that VLMs encode visual information correctly but fail to act on it---has direct implications for safety-critical deployments where faithful visual reporting is essential, and we hope it directs attention toward the arbitration mechanisms that need improvement.
 
We used an LLM assistant for limited rewriting and paraphrasing of parts of the paper; all experiments, analyses, and final claims in this manuscript were produced and verified by the authors.

\bibliography{references}
\bibliographystyle{colm2026_conference}

\appendix

\section{Implementation details}
\label{app:implementation}

All experiments use \texttt{float16}/\texttt{bfloat16} precision with \texttt{device\_map="auto"} across up to 4$\times$H200 GPUs. Input images are capped at 1024$\times$1024 pixels (aspect-ratio-preserving) before being passed to each model's vision encoder, which further processes them at its native resolution (e.g., CLIP-ViT-L/14: 336$\times$336 for LLaVA-v1.5; SigLIP-SO400M for LLaVA-OneVision; InternViT-6B with dynamic resolution for InternVL2). 

Crucially, our core finding---that models \emph{already encode} visual information correctly---means that higher input resolution would only \emph{strengthen} encoding, reinforcing the dissociation rather than resolving it; the bottleneck is downstream arbitration, not perceptual fidelity. For the 6-variant token matching protocol, we encode: (1) lowercase (``blue''), (2) capitalized (``Blue''), (3) uppercase (``BLUE''), (4) space-prefixed lowercase (`` blue''), (5) space-prefixed capitalized (`` Blue''), and (6) hex code subwords when available. The maximum logit across all variants is taken at each layer. This comprehensive matching is critical: BLIP-2's visual token rank improves from 170 (single-variant) to 2.8 (6-variant), correctly identifying it as a rank-based grounder.

\section{Paraphrase robustness details}
\label{app:paraphrase}

We test 8 question variants for color: (Q0) ``What color is the \{object\}?'' (baseline), (Q1) ``What color is this \{object\}?'', (Q2) ``Can you tell me the color of the \{object\}?'', (Q3) ``Describe the color of the \{object\} in the image.'', (Q4) ``The \{object\} in this image is what color?'', (Q5) ``Looking at the image, what color would you say the \{object\} is?'', (Q6) ``Identify the color of the \{object\}.'', (Q7) ``What is the color of the \{object\} shown here?'' For each model, we compute 100 samples across all 8 phrasings (800 forward passes per model).

\paragraph{Primary models.} Logit trajectory correlation: $r = 0.958$ (LLaVA-v1.5), $r = 0.961$ (LLaVA-Next), $r = 0.954$ (InternVL2). Winner agreement with baseline: 93.9\% (mean). Full stability (all 8 agree): 81.7\% (mean). Q7 agreement with Q0 remains high (88--97\%), confirming robustness.

\paragraph{Scaled-up models.} Qwen2-VL-72B maintains high rephrase stability (Q1--Q7 agreement: 89--95\%, mean 93\%), comparable to primary models. InternVL-26B is less stable (Q1--Q7 agreement: 67--92\%, mean 80\%); Q2 (``Can you tell me\ldots'') and Q3 (``Describe\ldots'') drop to 55--57\% visual success, while other phrasings remain above 76\%. This mid-range sensitivity suggests InternVL-26B's grounding is more dependent on question syntax than other models.

\paragraph{Size attribute.} We add reverse polarity (Q7: ``Which is \emph{smaller}?''), which yields 48--54\% agreement---near chance. This reveals keyword pattern-matching rather than genuine comparative reasoning.

\section{Additional MAC figures}
\label{app:mac-figures}

\subsection{Full MAC logit trajectories (all models)}
\label{app:mac-figures-full}

\begin{figure}[h]
\centering
\includegraphics[width=0.85\textwidth]{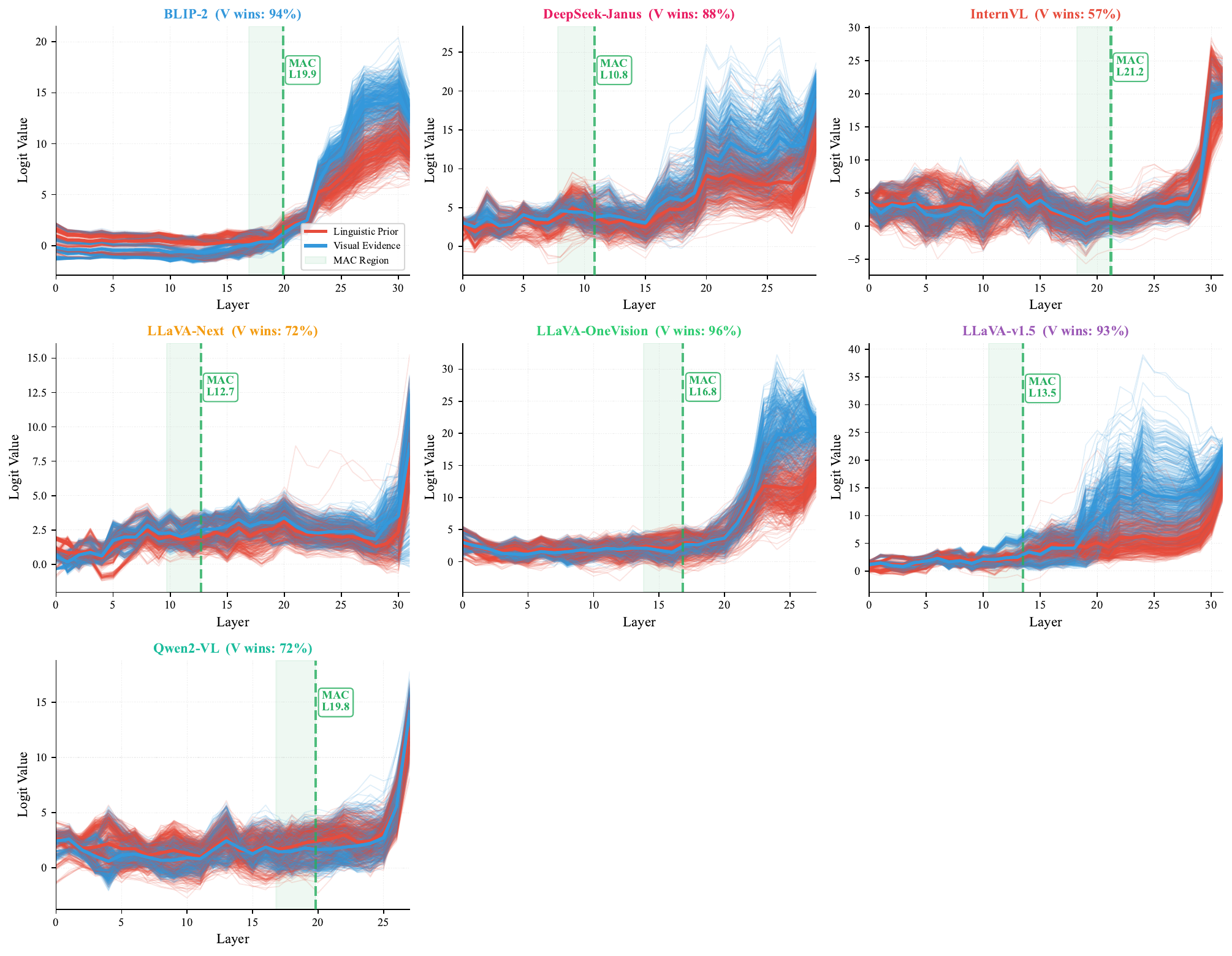}\\[4pt]
\includegraphics[width=0.85\textwidth]{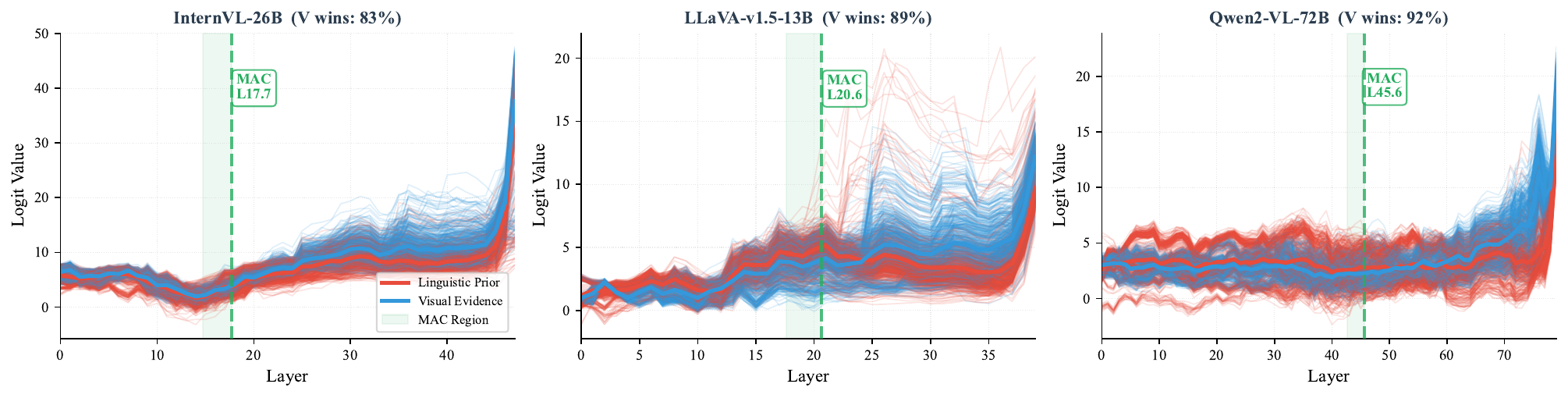}
\caption{MAC logit trajectories for all models (color, 493 samples). Each subplot shows per-layer visual (blue) and prior (orange) logits with sample traces and mean $\pm$1\,std shading. Green dashed line: MAC crossover. "V wins" indicates the percentage of samples where the visual logit exceeds the prior logit at the final layer. \textbf{Top:} seven primary models (7--8B). \textbf{Bottom:} three scaled-up variants (13--72B).}
\label{fig:mac-trajectories-full}
\end{figure}

\Cref{fig:mac-trajectories-full} shows MAC logit trajectories for all seven primary models and all three scaled-up variants.

\begin{figure}[h]
\centering
\includegraphics[width=0.9\textwidth]{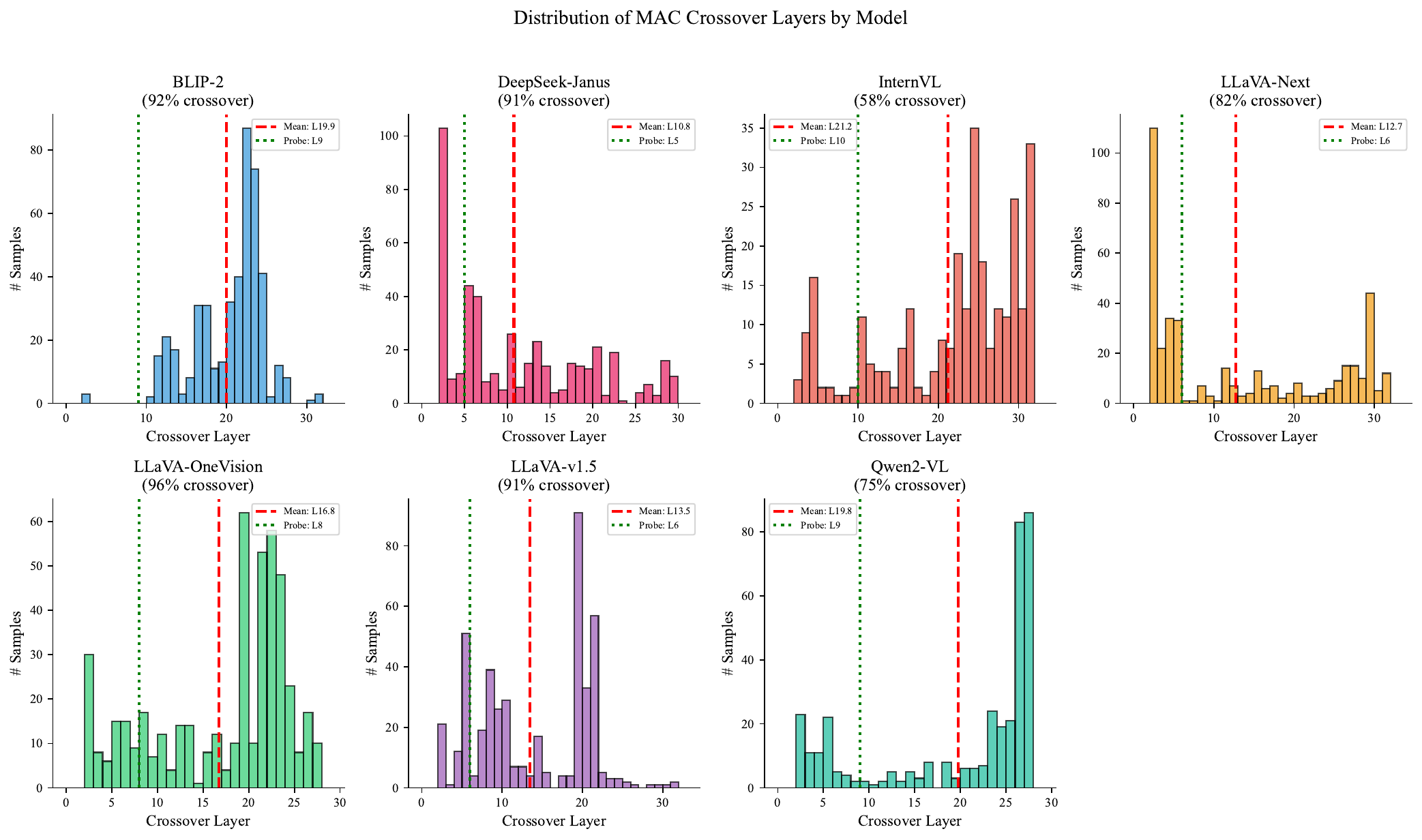}
\caption{Distribution of per-sample MAC layers across seven primary models (Color, 493 samples each). Box widths reflect crossover variance; fast integrators (DeepSeek-Janus, LLaVA-Next) show early medians with moderate spread, while slow integrators (Qwen2-VL, InternVL2) cluster near the final layers with tight distributions.}
\label{fig:mac-crossover-dist}
\end{figure}

\Cref{fig:mac-crossover-dist} shows the distribution of per-sample crossover layers across the seven primary models, revealing both the central tendency and variance of visual integration timing.

\section{Additional encoding and dissociation analysis}
\label{app:lt-fine}

\subsection{Fine-grained latent truth probing}

\begin{figure}[h]
\centering
\includegraphics[width=0.9\textwidth]{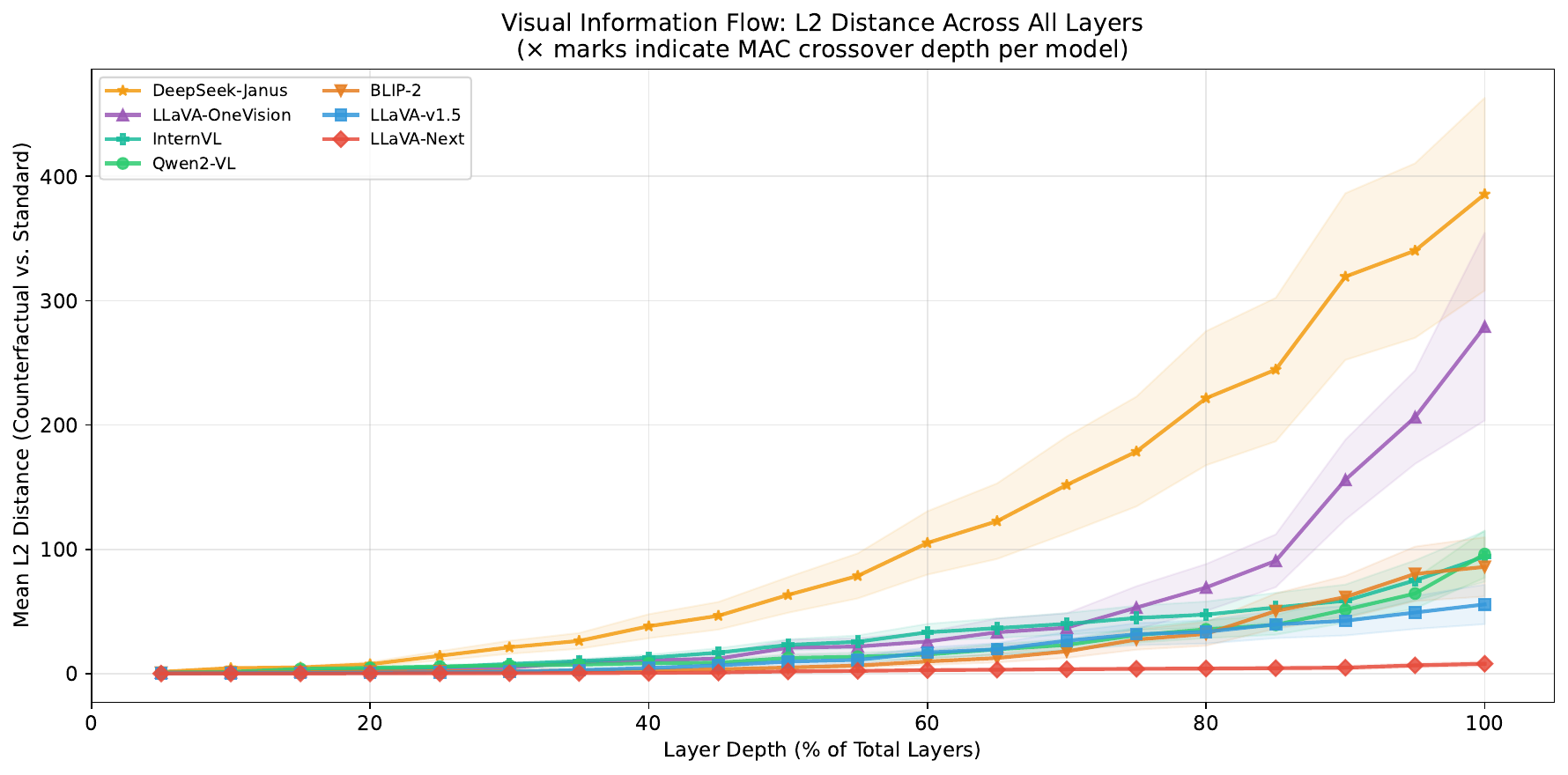}
\caption{Fine-grained visual encoding trajectory: L2 distance (counterfactual vs.\ standard) at 20 normalized depth points (5\%--100\%) across all seven primary models ($\times$ markers indicate MAC depth). Encoding grows monotonically and is never suppressed, reaching peak values at the final layer---the exact point where grounding failures occur.}
\label{fig:lt-fine}
\end{figure}

Fine-grained probing at 20 depth points (5\% to 100\% of total layers) across all 493 samples and 10 models confirms that L2 distance grows monotonically to the final layer (\Cref{fig:lt-fine}). Peak values at 100\% depth for primary models: DeepSeek-Janus (385.69), LLaVA-OneVision (279.10), Qwen2-VL (96.14), InternVL2 (94.13), BLIP-2 (85.83), LLaVA-v1.5 (55.65), LLaVA-Next (7.87). For scaled-up variants: Qwen2-VL-72B (700.61), InternVL2-26B (324.57), LLaVA-v1.5-13B (50.96). The 3-tier encoding hierarchy (strong/moderate/weak) is preserved at every depth, with larger models amplifying encoding strength by $2$--$5\times$.

\begin{figure}[h]
\centering
\includegraphics[width=0.9\textwidth]{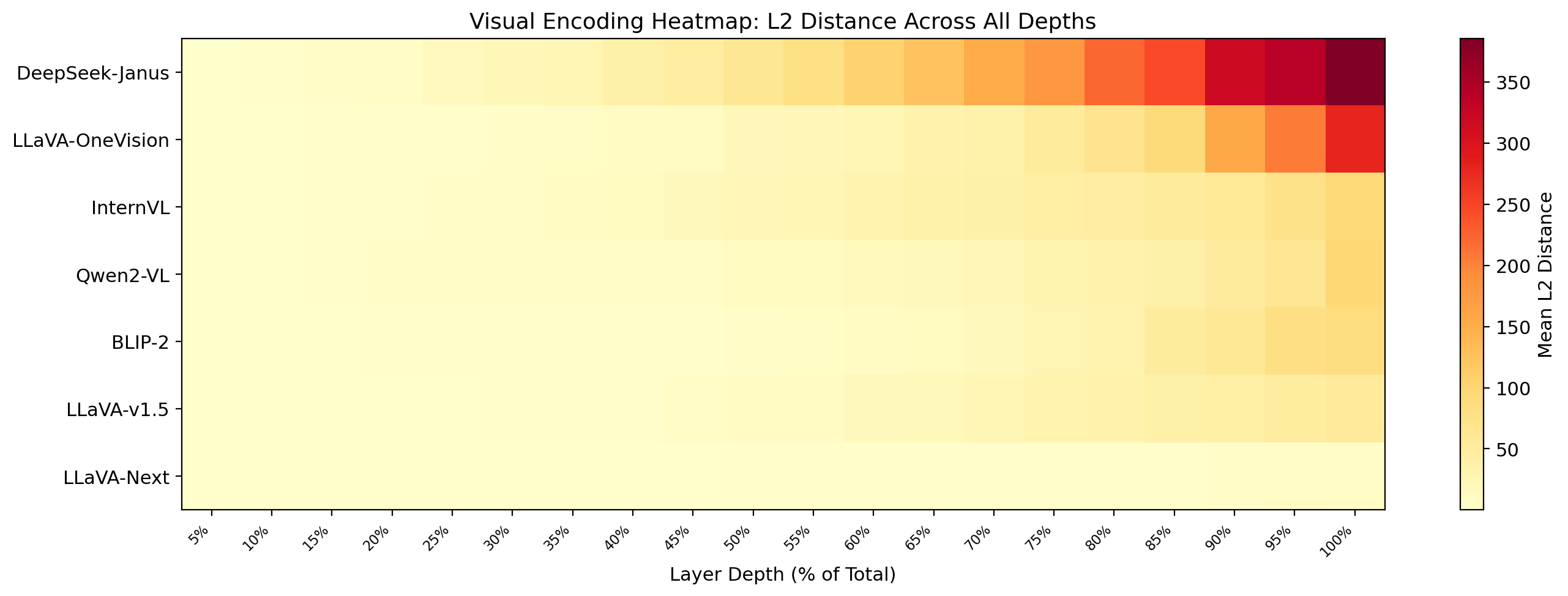}
\caption{Visual encoding heatmap: mean L2 distance across all depth points and seven primary models. Warmer colors indicate stronger visual encoding. DeepSeek-Janus and LLaVA-OneVision form the ``strong'' tier, InternVL2/Qwen2-VL/BLIP-2 the ``moderate'' tier, and LLaVA-v1.5/LLaVA-Next the ``weak'' tier.}
\label{fig:lt-heatmap}
\end{figure}

\Cref{fig:lt-heatmap} provides a complementary view as a heatmap, making the 3-tier encoding hierarchy visually apparent.

\subsection{Cross-stage correlation}
\label{app:cross-stage}

\begin{figure}[h]
\centering
\includegraphics[width=0.95\textwidth]{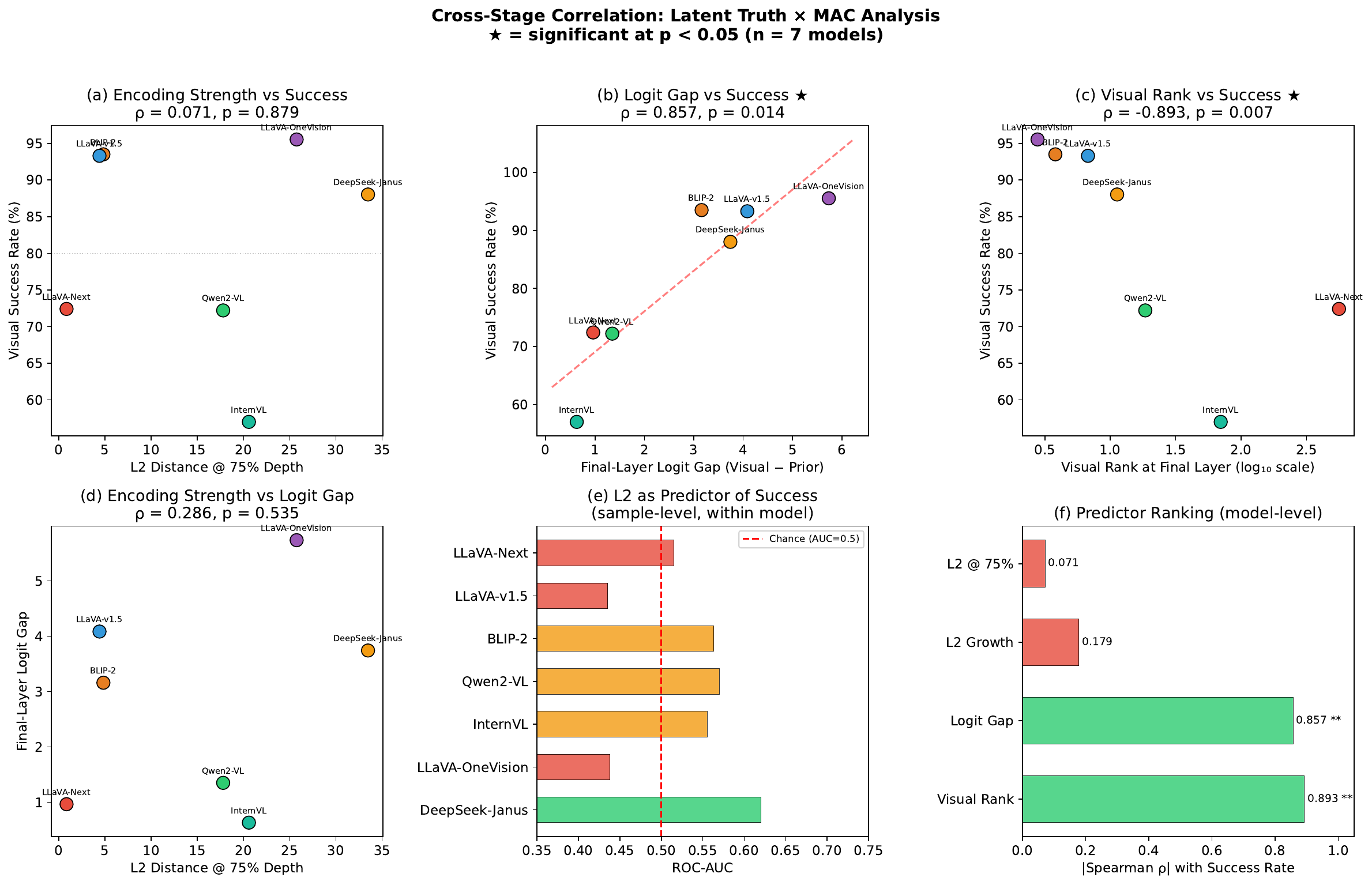}
\caption{Cross-stage correlation analysis ($n = 7$ primary models). (a)~Encoding strength (L2 at 75\% depth) shows no correlation with success ($\rho = 0.198$). (b)~Logit gap is strongly predictive ($\rho = 0.847$, $p = 0.016$). (c)~Visual rank is also predictive ($\rho = -0.893$). (d)~Encoding does not predict logit gap ($\rho = 0.464$). (e)~Sample-level ROC-AUC near chance (0.528). (f)~Arbitration-phase metrics predict success; encoding-phase metrics do not.}
\label{fig:cross-stage}
\end{figure}

\subsection{Success vs.\ failure encoding}
\label{app:lt-sf}

\begin{table}[h]
\centering

\small
\renewcommand{\arraystretch}{0.88}
\begin{tabular}{l|ccc|cc|r}
\toprule
& \multicolumn{3}{c|}{\textbf{Probe AUC}} & \multicolumn{2}{c|}{\textbf{$P(\text{visual})$ @ 10\%}} & \\
\textbf{Model} & \textbf{$\sim$10\%} & \textbf{$\sim$50\%} & \textbf{$\sim$97\%} & \textbf{Succ} & \textbf{Fail} & $\Delta$ \\
\midrule
InternVL2 & 1.000 & 1.000 & 1.000 & .958 & .961 & $-$.003 \\
DeepSeek-Janus & 1.000 & 1.000 & .998 & .990 & .991 & $-$.001 \\
LLaVA-OneVision & .995 & 1.000 & .998 & .892 & .915 & $-$.022 \\
Qwen2-VL & .995 & .999 & .999 & .905 & .909 & $-$.004 \\
BLIP-2 & .976 & .990 & .992 & .812 & .797 & $+$.015 \\
LLaVA-v1.5 & .939 & .987 & .975 & .653 & .650 & $+$.003 \\
LLaVA-Next & .865 & .991 & .988 & .517 & .521 & $-$.004 \\
\bottomrule
\end{tabular}
\caption{Linear probe AUC (5-fold CV, 493 samples) for decoding visual attributes from hidden states at three depths. Right columns: mean probe confidence for success vs.\ failure samples at $\sim$10\% depth.}
\label{tab:linear-probe}
\end{table}

\begin{figure}[h]
\centering
\includegraphics[width=0.9\textwidth]{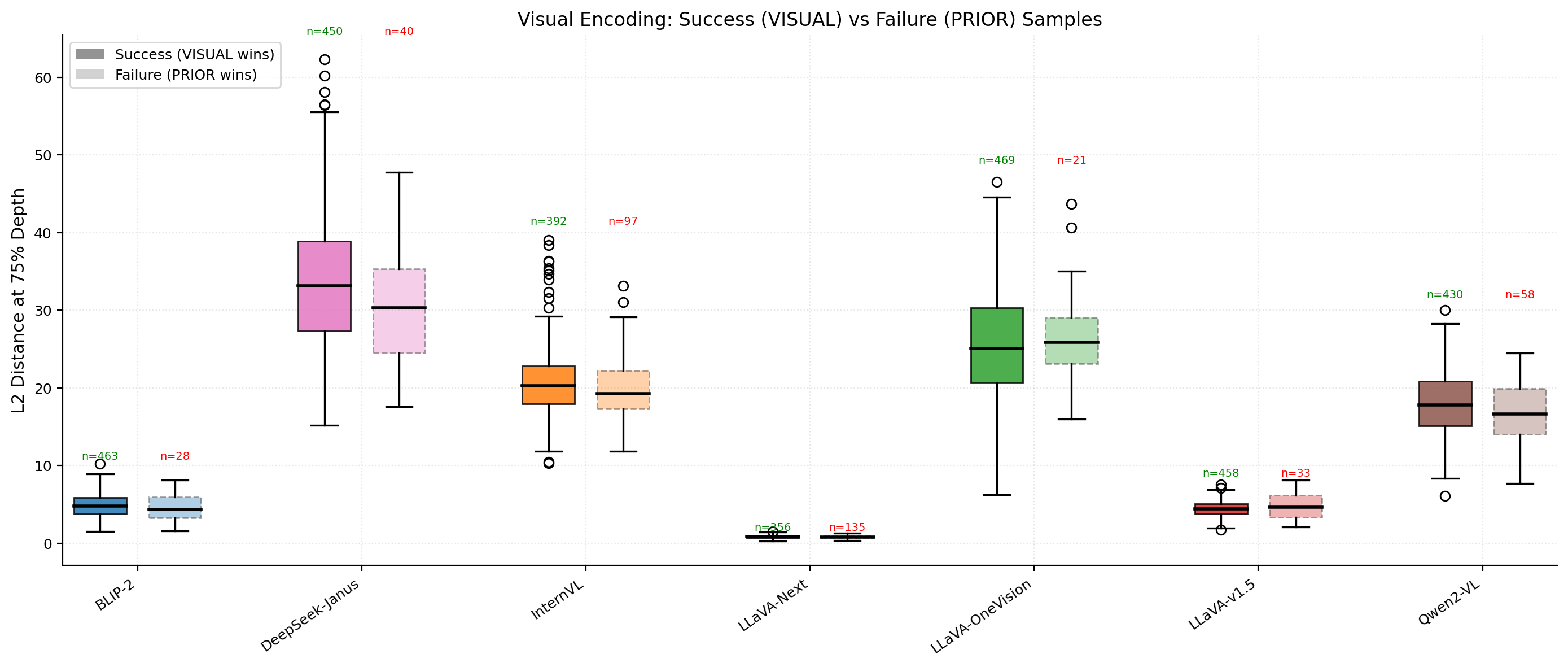}
\caption{Encoding strength in success vs.\ failure cases (L2 distance at 75\% MAC depth, Color). Dark boxes: samples where the model followed visual evidence (success). Light boxes: samples where the prior won (failure). Distributions overlap substantially for all seven models, with ratios in $[0.91, 1.12]$. Sample sizes shown in green (success) and red (failure).}
\label{fig:lt-sf-boxplot}
\end{figure}

\Cref{fig:lt-sf-boxplot} visualizes the encoding--grounding dissociation at the sample level: for each model, the L2 distributions of success (VISUAL wins) and failure (PRIOR wins) cases largely overlap, confirming that encoding strength does not predict grounding outcome.

\section{Additional patching analysis}
\label{app:patching-extra}

\begin{figure}[h]
\centering
\includegraphics[width=\textwidth]{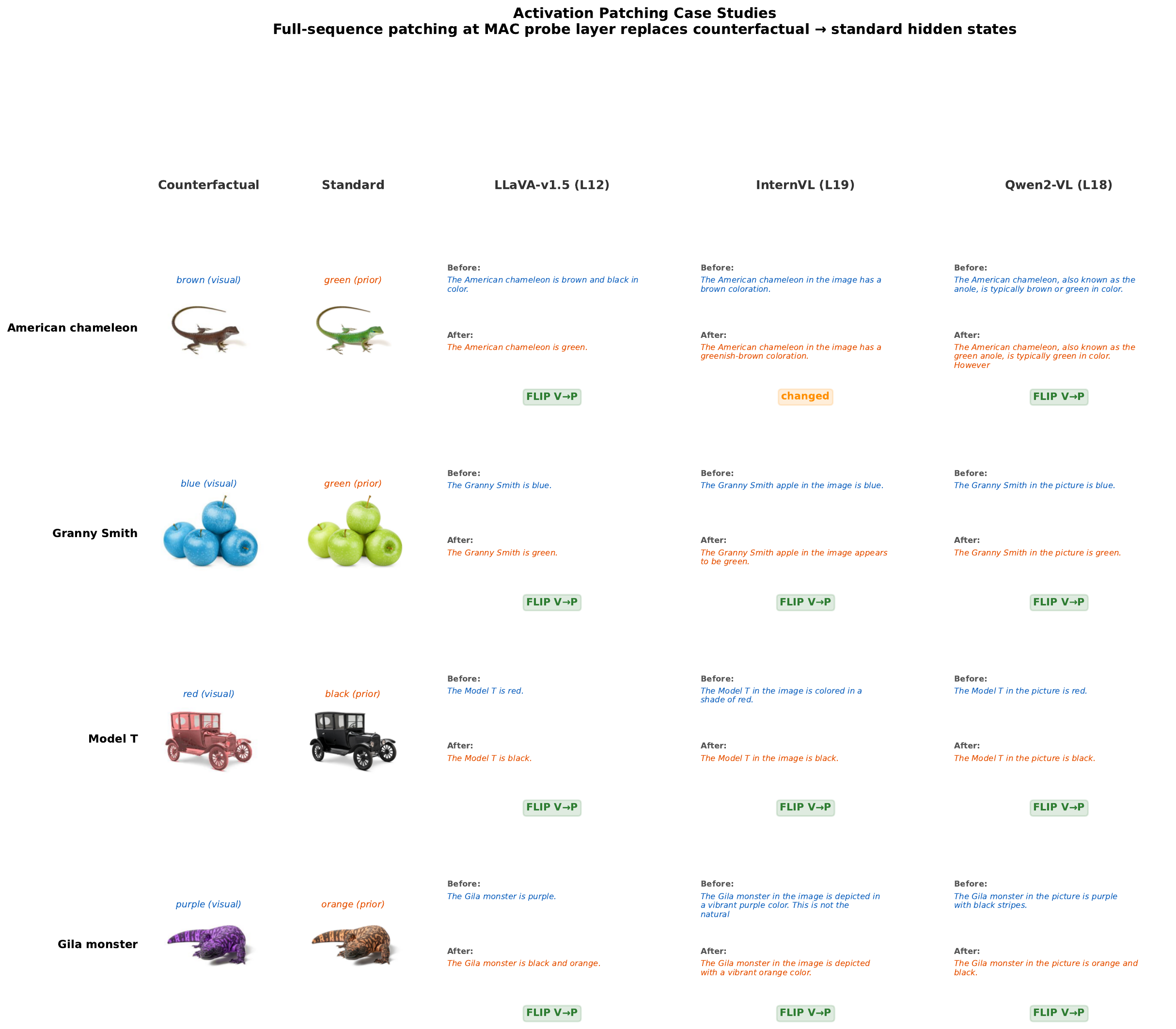}
\caption{Activation patching case studies. Each row shows a counterfactual image (altered color) and its standard counterpart, followed by baseline and patched outputs for three models. Full-sequence patching causally flips outputs from visual to prior answer.}
\label{fig:patching-cases}
\end{figure}

\begin{figure}[h]
\centering
\includegraphics[width=0.8\textwidth]{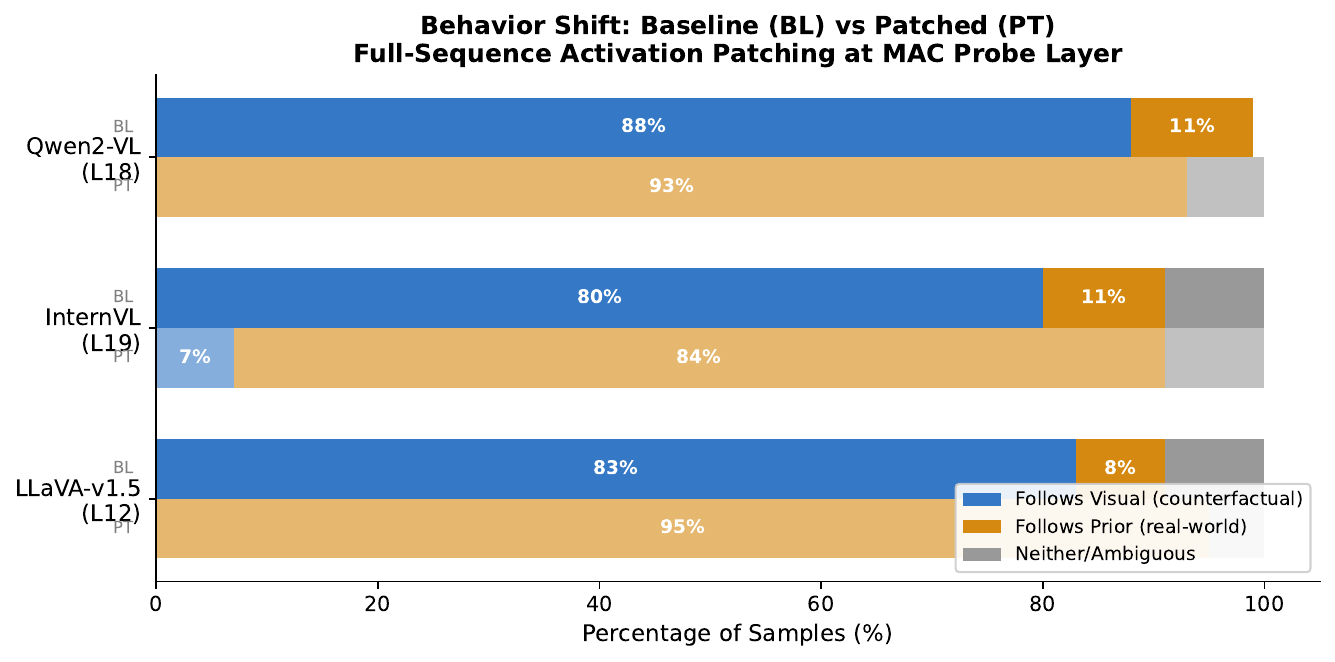}
\caption{Behavioral shift: baseline (BL) vs.\ patched (PT) output distributions. Blue: follows visual answer. Orange: follows prior answer. Full-sequence patching reverses the dominant behavior in all three models.}
\label{fig:patching-behavior}
\end{figure}

\Cref{fig:patching-behavior} shows the behavioral shift induced by full-sequence patching.

\section{Steering methodology details}
\label{app:steering-details}

This appendix provides the full technical details of the two steering methods introduced in \Cref{sec:intervention}.

\subsection{Linear steering}
\label{app:linear-steering-method}

Linear steering follows the contrastive activation addition paradigm introduced for text-only LLMs by \citet{DBLP:journals/corr/abs-2310-01405} and \citet{li2023inferencetime}. Given a set of training samples, we collect hidden states at a target layer $\ell$ for two conditions: counterfactual images (where the visual attribute has been altered, e.g., a blue banana) and standard images (the unmodified originals, e.g., a yellow banana). For each sample we run a forward pass up to layer $\ell$ and mean-pool the hidden state across all token positions to obtain a single vector $h_i \in \mathbb{R}^{d}$. The steering direction is:
\begin{equation}
    \mathbf{d} = \frac{1}{N_{\text{cf}}} \sum_{i=1}^{N_{\text{cf}}} h_i^{(\text{cf})} \;-\; \frac{1}{N_{\text{std}}} \sum_{i=1}^{N_{\text{std}}} h_i^{(\text{std})}
\end{equation}
This is the standard contrastive mean difference used for steering honesty~\citep{li2023inferencetime}, sentiment~\citep{turner2023steering}, and general behavioral control~\citep{DBLP:journals/corr/abs-2310-01405} in text-only LLMs. Our adaptation to VLMs differs in one critical respect: at inference, a hook at layer $\ell$ adds $\alpha \cdot \mathbf{d}$ to the hidden state at \emph{every} token position (not just the last token), and this hook remains active through all autoregressive generation steps. Full-sequence application is necessary because visual information is distributed across hundreds of image tokens (\Cref{sec:patching}). The strength $\alpha$ is swept over $\{0.0, 0.2, 0.5, 1.0, 1.5, 2.0, 3.0\}$.

\subsection{SAE-guided residual steering}
\label{app:sae-steering-method}

\paragraph{Step 1: SAE training.}
We train a one-hidden-layer Sparse Autoencoder~\citep{bricken2023monosemanticity, huben2024sparse} on the same hidden states used for linear steering. The architecture follows the standard dictionary learning formulation~\citep{bricken2023monosemanticity}: the SAE has encoder $E: \mathbb{R}^d \to \mathbb{R}^{d_{\text{sae}}}$ and decoder $D: \mathbb{R}^{d_{\text{sae}}} \to \mathbb{R}^d$, with $d_{\text{sae}} = 4d$ (expansion factor 4). The encoder uses ReLU activation to enforce non-negative, sparse feature activations:
\begin{equation}
    z = \text{ReLU}(W_{\text{enc}} \cdot h + b_{\text{enc}}), \qquad \hat{h} = W_{\text{dec}} \cdot z + b_{\text{dec}}
\end{equation}
Training minimizes reconstruction loss plus an L1 sparsity penalty~\citep{huben2024sparse}:
\begin{equation}
    \mathcal{L} = \| h - \hat{h} \|_2^2 + \lambda \| z \|_1, \qquad \lambda = 0.04
\end{equation}

\paragraph{Step 2: Feature selection.}
After training, we identify which SAE features correspond to visual evidence vs.\ linguistic priors. \citet{pach2025sparse} show that SAE features in VLMs learn monosemantic, interpretable directions; we leverage this by scoring features according to how differentially they activate across conditions. For each feature $j$, we compute:
\begin{equation}
    \Delta_j = \bar{z}_j^{(\text{cf})} - \bar{z}_j^{(\text{std})}
\end{equation}
where $\bar{z}_j^{(\text{cf})}$ and $\bar{z}_j^{(\text{std})}$ are the mean activations of feature $j$ across counterfactual and standard samples, respectively. Features are then scored by their decoder-weighted differential activation:
\begin{equation}
    s_j = |\Delta_j| \times \| W_{\text{dec}}[:,j] \|_2
\end{equation}
The decoder-norm weighting accounts for the fact that different features may have different magnitudes of effect on the output space~\citep{huben2024sparse}. The top-50 features with $\Delta_j > 0$ (more active for counterfactual images) are classified as \emph{visual features}; the top-50 with $\Delta_j < 0$ are classified as \emph{prior features}.

\paragraph{Step 3: Residual steering.}
Standard SAE-based steering~\citep{templeton2024scaling} directly clamps or replaces feature activations and decodes the result, effectively substituting the hidden state with the SAE reconstruction $h'_t = \hat{h}'$. We adopt two adaptations for the VLM setting. First, we steer \emph{bidirectionally}---simultaneously amplifying visual features and suppressing prior features. Second, inspired by residual connections~\citep{he2016deep}, we apply a residual strategy that adds only the effect of the feature manipulation to the original hidden state, avoiding the information loss of full reconstruction. At inference, a hook at layer $\ell$ performs the following for each forward pass:
\begin{enumerate}[leftmargin=*,nosep]
    \item Mean-pool the hidden state across token positions: $\bar{h} = \frac{1}{T}\sum_t h_t$.
    \item Encode through the SAE: $z = \text{ReLU}(W_{\text{enc}} \cdot \bar{h} + b_{\text{enc}})$.
    \item Construct a steering vector in feature space:
    \begin{equation}
        z'_j = \begin{cases}
            z_j + \alpha_v & \text{if } j \in \mathcal{F}_{\text{visual}} \\
            z_j - \alpha_p & \text{if } j \in \mathcal{F}_{\text{prior}} \\
            z_j & \text{otherwise}
        \end{cases}
    \end{equation}
    where $\alpha_v, \alpha_p \geq 0$ are the visual and prior steering strengths, swept over $\{0, 1, 2, 3, 5\}$. All values are clamped to $\geq 0$ after modification.
    \item Decode both the original and modified feature vectors:
    \begin{equation}
        \hat{h} = D(z), \qquad \hat{h}' = D(z')
    \end{equation}
    \item Compute the delta and add it to the \textbf{original} hidden state at every token position:
    \begin{equation}
        h'_t = h_t + (\hat{h}' - \hat{h}), \qquad \forall\, t \in \{1, \ldots, T\}
    \end{equation}
\end{enumerate}

\paragraph{Why residual application?} The standard approach~\citep{templeton2024scaling} replaces the hidden state with the modified SAE reconstruction: $h'_t = \hat{h}'$. This discards all information the SAE did not learn to reconstruct. \citet{bricken2023monosemanticity} note that SAE reconstructions are lossy, particularly for features outside the training distribution. The residual formulation isolates the \emph{effect} of the feature manipulation ($\hat{h}' - \hat{h}$) and applies it additively to the original hidden state, preserving all information not captured by the SAE---analogous to how residual connections~\citep{he2016deep} preserve information across transformer layers. This design choice is critical: in pilot experiments, direct replacement degraded baseline accuracy by 2--5\%, whereas the residual formulation maintained it exactly.

\end{document}